\pdfoutput=1
\documentclass[10pt]{article}
\usepackage[preprint]{tmlr} %
\usepackage[utf8]{inputenc}
\usepackage{amsmath,amssymb,amsthm,mathtools}
\usepackage{booktabs}
\usepackage{tabularx}
\usepackage{graphicx}
\usepackage{hyperref}
\usepackage[capitalise,noabbrev]{cleveref} %

\usepackage{aliascnt}
\newcommand{\sharedtheorem}[2]{\newaliascnt{#1}{proposition}\newtheorem{#1}[#1]{#2}\aliascntresetthe{#1}}
\newtheorem{theorem}{Theorem}
\newtheorem{proposition}{Proposition}
\sharedtheorem{lemma}{Lemma}
\sharedtheorem{corollary}{Corollary}
\theoremstyle{definition}
\newtheorem{definition}{Definition}
\sharedtheorem{appdefinition}{Definition}
\crefname{appdefinition}{Definition}{Definitions}
\newtheorem{example}{Example}
\theoremstyle{remark}
\sharedtheorem{remark}{Remark}
\usepackage{enumitem}
\usepackage{etoolbox}
\newcommand{\alphitems}{\setlist[enumerate,1]{label=(\alph*),ref=(\alph*)}}
\AtBeginEnvironment{theorem}{\alphitems}\AtBeginEnvironment{proposition}{\alphitems}
\AtBeginEnvironment{lemma}{\alphitems}\AtBeginEnvironment{corollary}{\alphitems}
\AtBeginEnvironment{definition}{\alphitems}\AtBeginEnvironment{example}{\alphitems}
\AtBeginEnvironment{remark}{\alphitems}\AtBeginEnvironment{appdefinition}{\alphitems}
\newcommand{\appendixnumbering}{%
  \crefalias{section}{appendix}%
  \setcounter{proposition}{0}\counterwithin*{proposition}{section}%
  \renewcommand{\theproposition}{\thesection.\arabic{proposition}}%
  \renewcommand{\theHproposition}{app.\thesection.\arabic{proposition}}%
  \aliascntresetthe{lemma}\aliascntresetthe{corollary}\aliascntresetthe{remark}\aliascntresetthe{appdefinition}%
  \def\theHlemma{app.\thesection.lem.\arabic{proposition}}%
  \def\theHcorollary{app.\thesection.cor.\arabic{proposition}}%
  \def\theHremark{app.\thesection.rem.\arabic{proposition}}%
  \def\theHappdefinition{app.\thesection.def.\arabic{proposition}}}

\newcommand{\E}{\mathbb{E}}
\newcommand{\R}{\mathbb{R}}
\newcommand{\Cov}{\operatorname{Cov}}
\newcommand{\tr}{\operatorname{tr}}
\newcommand{\tail}{\operatorname{tail}}
\newcommand{\row}{\operatorname{row}}
\newcommand{\ran}{\operatorname{ran}}
\newcommand{\Proj}{\operatorname{Proj}}
\newcommand{\aff}{\mathrm{aff}}
\newcommand{\Dlin}{D_{\mathrm{lin}}}
\newcommand{\Kc}{\mathcal{K}}
\newcommand{\Qc}{\mathcal{Q}}
\newcommand{\Hc}{\mathcal{H}}
\newcommand{\Fc}{\mathcal{F}}

\newcommand{\Wc}{\mathcal{W}}
\newcommand{\Ec}{\mathcal{E}}
\newcommand{\zh}{\hat z}
\newcommand{\cbar}{\bar c_t}
\newcommand{\Abar}{\bar A}
\newcommand{\Rbar}{\bar R}

\title{Information Is Not Implementation:\\ Known Structure and Generative Learnability\\ in Visual Latent Spaces}
\author{\name Shiyu Chen \email cyrilcsy@gmail.com \\
\addr Hosei University
\AND
\name Jinjia Zhou \email zhou@hosei.ac.jp \\
\addr Hosei University}

\begin{document}
\maketitle

\begin{abstract}
Diffusion and flow models on high-dimensional representation latents can fail with velocity or noise targets when network width is comparable to token dimension, and each of a growing set of fixes has its own explanation. We develop a unified theoretical perspective on the generative learnability of visual latent spaces that separates what these explanations conflate: the source of a learning burden, the information available about the representation, and the operation that exploits it. Measuring burden by the reducible error of an output-rank-limited learner shows that information is not implementation: on low-rank latents with data and noise on the same sphere, adding the network output to an exact radial compensation leaves positive target variance in every direction the data do not occupy, whereas projecting the output with the same compensation admits an optimal output within the data subspace and costs nothing once the output rank accommodates the data. Three relations organise the perspective. First, clean-data prediction, preconditioning, tangent projection and learned skips act on one burden, the pass-through term proportional to the input; the norm constraint does not create it but makes its removal lossless. Second, the scalar offload that standard preconditioning chooses by total target variance carries almost the same radial information as the exact compensation and leaves almost the same variance in the directions the data do not occupy, since an output-rank-limited learner pays for the tail of the target's spectrum, not its trace. Third, second-order statistics and an exact constraint determine overlapping field components for any affine constraint, such as a subspace, whereas a constraint whose direction varies with the state, such as a fixed norm, can determine complementary ones, as measured on representation-autoencoder latents. We derive principles for capacity-aware parameterisation, statistical--geometric hybrid formulations and decoder-aware allocation, the last because a component's learning difficulty differs from the consequence of its errors after decoding: the mean-squared local response of the DINOv2 and SigLIP2 decoders to an error in the directions the latents barely occupy is about an order of magnitude larger than to one of the same energy in the principal directions.
\end{abstract}

\section{Introduction}\label{sec:intro}

In latent generative models for images, a tokenizer encodes images into a grid of latent tokens, a diffusion or flow model learns to generate such grids, and the tokenizer's decoder turns them back into images. Recent tokenizers built on pretrained representation encoders produce tokens of several hundred dimensions \citep{rae}, and on such latents standard diffusion targets can fail when the network width is comparable to the token dimension. With DINOv2-B latents of dimension $768$ and a transformer of width $768$, Euclidean flow matching reaches FID $24.21$, against $6.77$ for a Riemannian variant \citep{rjf}; %
on $16\times16$ pixel patches of the same dimension, velocity prediction fails where clean-data prediction succeeds \citep{jit}. %

\begin{figure}[t]%
\centering
\includegraphics[width=\linewidth]{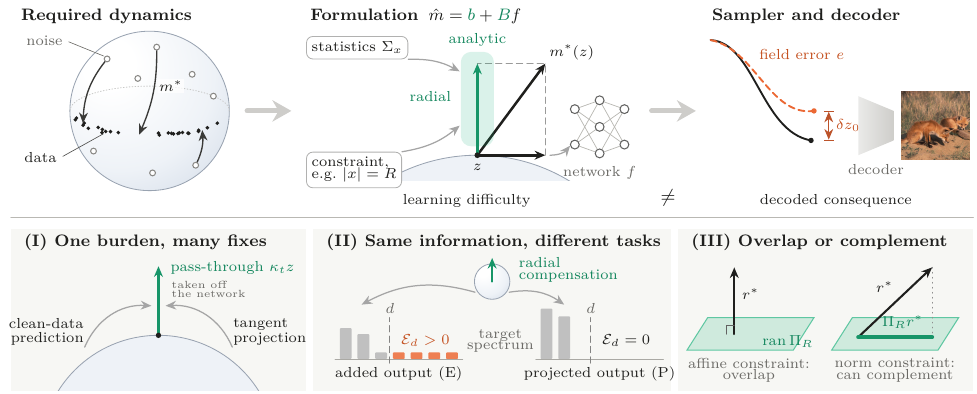}
\caption{Overview (schematic). A representation shapes learnability twice: its distribution fixes the required dynamics $m^*$, and its structure, here $\Sigma_x$ and $|x|=R$, lets a formulation $\hat m=b+Bf$ realise part of $m^*$ analytically (green), leaving the rest to the network~$f$, whose field error then passes through the sampler and decoder. (I)~Clean-data prediction and the tangent projection, built on different information, take the same pass-through term $\kappa_tz$ off the network (\cref{thm:relation-i}). (II)~With the same exact radial compensation, an added output leaves target variance beyond the output rank $d$ and a projected one none (bars: spectrum of the network's target, for (P) of its optimal output $D$, drawn for data of rank $r$ and $r\le d<n-r$; $\Ec_d$: reducible error; \cref{thm:relation-ii}). (III)~Statistics and a constraint overlap when the constraint determines nothing beyond the affine prediction, $\Pi_Rr^*=0$ with $r^*=m^*-a^*$, and can complement each other otherwise (\cref{thm:relation-iii}).}%
\label{fig:overview}
\end{figure}

Several fixes are now available, each with its own explanation: tangent projections on the hypersphere, motivated by geometric interference \citep{rjf,aligning}; clean-data prediction, motivated by the manifold hypothesis or a collapse of effective dimension \citep{jit,diffusibility,rit}; wider networks, heads or residual streams \citep{rae,rsb}; and preconditioning and mixtures of prediction targets adapted to the data statistics \citep{edm,jinwang,hybridspectral}. \Citet{rsb} have shown that parameterisation and architectural fixes work for a common reason. Still, how the constraint-based fixes relate to the statistical and parameterisation-based ones, and what it is about a representation that lets any of them work, has no common mathematical description. In practice, one cannot tell for a new representation or network which mechanism matters, and comparisons of representations depend on formulation choices that are rarely stated.

The gap shows in a simple question. %
With data and noise on the same sphere and the chord path, the radial part of the required dynamics is a known function of the input. Suppose two formulations supply exactly this radial part analytically and differ only in how the network output enters the field: added to it, or first projected onto the tangent space. Does the network have the same thing to learn? It does not. The added output has a single optimal value, the tangential part of the denoiser $D=\E[x\mid z]$, whereas the projected output only has to fall in a set of equally good outputs, a set that contains $D$ itself (\cref{prop:fiber}). On low-rank data that single value has variance in every direction the data do not occupy, whereas $D$ stays within the data subspace, so a network whose outputs span as many dimensions as the data can realise the projected formulation exactly, but not the added one (\cref{thm:relation-ii}). The same information can leave a finite network different tasks.

We develop a unified theoretical perspective on visual latent generative learnability built on this separation. A representation shapes learnability twice: its distribution fixes the dynamics to be realised, and its statistical and geometric structure determines which parts of them can be realised analytically. The dynamics contain the sources of the learning burden, the structure is the information available to the generative system, and a formulation adds the operations that exploit that information; the explanations above run these three things together. A burden is a part of the network's task that occupies a finite network's capacity; below the token dimension we measure it by the reducible error of an output-rank-limited learner (\cref{sec:measure-burden}). At width equal to the token dimension, as in both examples above, output rank limits nothing, and the part of the target proportional to the input is paid for in residual-stream bandwidth instead (\cref{sec:reading-i}). %
Within this view we establish three relations (\cref{fig:overview}; \cref{thm:relation-i,thm:relation-ii,thm:relation-iii}): Relation~I identifies the burden that the published fixes share; Relation~II fixes the known information and varies the operation, of which the question above is an instance; and Relation~III asks what a geometric constraint adds to the second-order statistics. \Cref{sec:decoded} then separates the difficulty of learning a component from the consequence of its errors after decoding, completing the line from required dynamics to images.

\paragraph{Contributions.}\leavevmode
\begin{enumerate}
\item \emph{Foundational understanding of generative learnability.} Learnability depends not only on what is known about a representation but on how that knowledge is implemented: the offload that minimises the total variance of the network's target, as standard preconditioning does, need not be the best one for a network of limited capacity (\cref{thm:relation-ii}). Statistical and geometric knowledge overlap for affine constraints such as a subspace and can complement each other for constraints whose direction varies with the state, such as a fixed norm (\cref{thm:relation-iii}); using more statistics does not imply greater consistency with the geometry (\cref{ex:two-clusters}), which we also observe on real representation latents (\cref{fig:relation-iii}). %
\item \emph{Mechanistic reinterpretation and unification of existing methods.} Clean-data prediction, preconditioning, tangent projection on the hypersphere and learned skip connections act on a common component of the target, the pass-through term, which is proportional to the input and therefore radial for any data (\cref{thm:relation-i}, \cref{tab:formulations}). This refines the geometric accounts of why spherical methods succeed \citep{rjf,rit}: the norm constraint does not create this burden; it is the information that allows the burden to be removed without loss. It also gives the manifold and effective-dimension accounts of clean-data prediction \citep{jit,diffusibility} a precise form: what clean-data prediction avoids is the pass-through cost in the directions the data do not occupy. %
\item \emph{Theoretical principles for generative system design} (\cref{sec:disc-design}): capacity-aware parameterisation, statistical--geometric hybrid formulations and decoder-aware allocation, the last because the decoders of representation autoencoders weight errors of the same size very differently by direction (\cref{fig:decoder}). %
\end{enumerate}

\section{Setting: required dynamics and known structure}\label{sec:setting}

We write $x\in\R^n$ for one latent token and $z$ for its noisy version, $\mathbf Q$ for the distribution of the whole clean grid and $Q$ for its marginal at a uniformly drawn position, with covariance $\Sigma_x$, eigenvalues $\lambda_1\ge\dots\ge\lambda_n$, and zero mean unless stated otherwise. Bayes quantities condition on the whole noisy grid, although we write $\E[\,\cdot\mid z]$, $D(z,t)$ and $m^*(z,t)$; known statistics are the first two moments of $Q$ (\cref{app:notation} gives the conventions).

\subsection{Paths, targets and the Bayes field}

We consider linear interpolation paths
\begin{equation}\label{eq:path}
  z=\alpha_t x+\sigma_t\varepsilon,\qquad t\in[0,1],
\end{equation}
with $t=0$ the data end; the chord path $\alpha_t=1-t$, $\sigma_t=t$ is the running example. The noise $\varepsilon$ is independent of $x$ with $\Cov(\varepsilon)=I$; it is either Gaussian or uniform on the sphere of radius $R=\sqrt n$, the spherical normalisation used by \citet{rjf}. Several works we discuss place the data at $t=1$; we convert their statements by $t\mapsto 1-t$. Substituting $\varepsilon=(z-\alpha_tx)/\sigma_t$, every regression target $U=a_tx+b_t\varepsilon$ can be written as%
\begin{equation}\label{eq:target}
  U=\kappa_t z+\beta_t x,\qquad \kappa_t=\frac{b_t}{\sigma_t},\quad \beta_t=a_t-\frac{b_t\alpha_t}{\sigma_t},\qquad m_U:=\E[U\mid z]=\kappa_t z+\beta_t D,
\end{equation}
with $D=\E[x\mid z]$. On the chord path, $(\kappa_t,\beta_t)$ equals $(1/t,\,-1/t)$ for the velocity $\varepsilon-x$, $(1/t,\,-(1-t)/t)$ for the noise, and $(0,1)$ for clean data. We call the Bayes field of the path velocity $v=\dot\alpha_tx+\dot\sigma_t\varepsilon$ the \emph{required dynamics} and write it $m^*:=m_v$. It is the field the sampler integrates; the fields $m_U$ of other targets are fixed affine transforms of it, and changing the path changes $m^*$. Unless stated otherwise, $\kappa_t$ and $\beta_t$ refer to the velocity. The data enter $m^*$ only through $D$, whereas the \emph{pass-through term} $\kappa_tz$ is a fixed function of the input. %
It exists for every data distribution and vanishes for clean-data prediction; it is a property of the target, not of the representation. We write $\Dlin$ for the linear minimum-mean-squared-error (LMMSE) estimator of $x$ from the token's own $z$, $a^*:=\kappa_t z+\beta_t\Dlin$ for the affine part of $m^*$, which is determined by the first two moments of the latent distribution, and $r^*:=m^*-a^*=\beta_t(D-\Dlin)$ for its residual beyond per-token affine prediction.

\subsection{Known information}\label{sec:known-information}

Besides samples, a generative system may know three kinds of facts about the latents: the path and the noise law, which it always knows; second-order statistics, such as the mean and covariance; and constraints satisfied by every latent, such as membership in a known subspace or a fixed norm. We collect these facts in $\Kc$ and write $\Qc_\Kc$ for the set of grid distributions compatible with them. Fields are elements of $\Hc=L^2(p_t;\R^n)$, where $p_t$ is the joint law of the noisy grid $Z$ and a uniformly drawn position $i$, with $z=Z_i$, and $\langle f,g\rangle=\E[f^\top g]$; $\Hc_\aff$ denotes the affine functions $z\mapsto Az+c$ of the token's own $z$, with $A$ and $c$ shared across positions, and $\Pi_\aff$ the orthogonal projection onto them. %

\begin{definition}[Known relations]\label{def:known}%
\leavevmode
\begin{enumerate}
\item \emph{Moment relation.} If the mean and covariance are known and $\Cov(z)$ is invertible, then $\Pi_\aff m^*=a^*$, and $a^*$ is the same for every $\mathbf Q\in\Qc_\Kc$ (orthogonality principle).
\item \emph{Pointwise relation.} Suppose there are $L(z)$ and $h(z)$, depending only on $(z,t)$, such that $L(z)\,v=h(z)$ almost surely for every $\mathbf Q\in\Qc_\Kc$. Then $Lm^*=h$. Let $\Pi_R$ act by pointwise multiplication with $L^\dagger L$ and set $P_C=I-\Pi_R$. The relation determines $\Pi_Rm^*=L^\dagger h$, and the fields that satisfy it are exactly $L^\dagger h+P_Cw$ for arbitrary $w$.
\end{enumerate}
\end{definition}

A moment relation holds on average over $p_t$; a pointwise relation holds at every $z$. Two kinds of pointwise relation recur, according to whether $L$ depends on $z$, and Relations~II and~III treat them differently.%

\paragraph{Affine constraint: subspace.} If $Lx=c$ almost surely with fixed $L$ and $c$, then $Lv=\kappa_tLz+\beta_tc$, a pointwise relation with fixed $L$. A known $r$-dimensional subspace $S$, a model of low effective dimension \citep{diffusibility}, gives $L=P_{S^\perp}$ and $h=\kappa_tP_{S^\perp}z$.

\paragraph{Quadratic constraint: sphere, chord path.} If $|x|=|\varepsilon|=R$ on the chord path, then every sample satisfies
\begin{equation}\label{eq:sphere-h}
  z^\top x=h_D(|z|^2):=\frac{|z|^2+(1-2t)R^2}{2(1-t)},\qquad z^\top v=h_m:=\kappa_t|z|^2+\beta_th_D=\frac{(2t-1)(R^2-|z|^2)}{2t(1-t)},
\end{equation}
so the radial component of the target is a known function of $z$ (\cref{thm:relation-i}). In \cref{def:known}, $L=z^\top$ and $h=h_m$ for the velocity, and $h_D$ for the denoiser; $P_C=P_T:=I-\zh\zh^\top$ with $\zh=z/|z|$ is the tangent projection used by \citet{rjf}. Here $L$ varies with $z$. The same holds on any linear path for a quadratic constraint $x^\top Ax=\varepsilon^\top A\varepsilon=c$, since then $z^\top Ax=\bigl(z^\top Az+(\alpha_t^2-\sigma_t^2)c\bigr)/(2\alpha_t)$; the sphere is the case $A=I$.%

Representation-autoencoder latents carry one constraint of each kind: the final LayerNorm, applied without affine parameters, gives every token zero mean over channels, $\mathbf 1^\top x=0$, and norm $\sqrt n$ (\cref{app:latents}).

\section{Formulations and the learner's task}\label{sec:formulations}

\begin{definition}[Formulation]\label{def:formulation}%
A \emph{formulation} implements the field as $\widehat m=b+Bf$, where $b(z,t)$ and $B(z,t)$ are fixed by $\Kc$ and the path and $f$ is the network output. It is \emph{Bayes-consistent} if $m^*-b$ lies pointwise in the range of $B$.
\end{definition}

Since $\E\|b+Bf-v\|^2=\E\|b+Bf-m^*\|^2+\E\|v-m^*\|^2$, training with the velocity loss is regression of $b+Bf$ onto $m^*$ in the field metric; losses stated in other units, such as clean-data or noise losses, differ only by a time weight, which we write as $q_t$. \Cref{tab:formulations} places representative methods in this form.

\begin{table}[t]
\centering
\footnotesize
\caption{Representative formulations $\widehat m=b+Bf$ (field units, $t=0$ the data end). ``Burden removed'' names the part of the network's target that the fixed operations take over (\cref{sec:measure-burden}); entries for the projection methods are proved in \cref{sec:relation-i}. Coefficients stated in other conventions are converted with \eqref{eq:target}.}
\label{tab:formulations}
\begin{tabularx}{\linewidth}{@{}>{\raggedright\arraybackslash}p{3.3cm}>{\raggedright\arraybackslash}X>{\raggedright\arraybackslash}p{1.9cm}>{\raggedright\arraybackslash}p{2.5cm}>{\raggedright\arraybackslash}X@{}}
\toprule
Method & $b$ & $B$ & Information used & Burden removed\\
\midrule
Velocity prediction & $0$ & $I$ & path & ---\\
Clean-data prediction \citep{jit,diffusibility,rit} & $\kappa_t z$ & $\beta_t I$ & path & pass-through\\
EDM preconditioning \citep{edm} & $(\kappa_t+\beta_t c_{\rm skip})z$ & $\beta_t c_{\rm out}I$ & scalar second moment & pass-through and scalar linear part, but hands back $-c_{\rm skip}z$ (\cref{thm:relation-ii})\\
Learned target mixing \citep{jinwang} & $\kappa^{(k)}_t z$ & $\gamma^{(k)}_t I$ & path; scalar $k$ learned from samples & part of pass-through\\
Learned skip \citep{rsb} & $\kappa(t)z$, learned & $\gamma(t)I$, learned & none & pass-through (learned)\\
Wiener mixture \citep{hybridspectral} & $a^*=\kappa_tz+\beta_t\Dlin$ & $\beta_t\sigma_tI$ & mean and covariance & full affine part\\
Tangent projection \citep{rjf} & $0$ (geodesic path) & $\propto P_T(z)$ & norm constraint & pass-through, all radial output\\
Projection, chord path & $L^\dagger h_m$ & $\beta_tP_T(z)$ & norm constraint & pass-through, all radial output\\
Normal contraction \citep{msfm} & $-\alpha(t)\,r(z)$ & $P_T(z)$ & estimated tangent space & none (sampling stability)\\
\bottomrule
\end{tabularx}
\end{table}

\subsection{What the network is asked to learn}

\begin{proposition}[The learner's target is a fiber]\label{prop:fiber}%
Over all measurable $f$, the optimal outputs form the set $\{f:\ Bf=\Pi_{\ran B}(m^*-b)\ \text{almost surely}\}$, and the formulation itself leaves the error $\E\|(I-\Pi_{\ran B})(m^*-b)\|^2$, which no learner can remove and which vanishes exactly when the formulation is Bayes-consistent.
\begin{enumerate}
\item \emph{Affine family.} If $b$ is affine in $z$ and $B=\gamma_tI$, the target is the single point $\gamma_t^{-1}[(a^*-b)+r^*]$, whose two parts are orthogonal.
\item \emph{Projection.} If $B=P_C$, the network's output along $\row L$ does not enter the field, so errors in those directions cost nothing. Every element of the fiber is optimal within the formulation, and the network may realise whichever element is easiest to represent.
\end{enumerate}
\end{proposition}

The proof is pointwise least squares (\cref{app:proofs}). A Bayes-consistent formulation leaves the required dynamics unchanged and changes only the task of the network. An affine offload translates the target; a projection does not subtract anything, but enlarges the target from a point to a set. Architectures and formulations enter the learning problem at different places. In terms of representability, the best field a class $\Fc$ can realise depends on $(\Fc,b,B)$ only through $\{b+Bf:f\in\Fc\}$, so an architecture raises what the network can represent, whereas a formulation changes what it must represent. %
Exact constraints can be added to any formulation without loss.

\begin{proposition}[Imposing an exact constraint]\label{prop:impose}%
Suppose $Lm^*=h$ holds exactly. For any formulation $\widehat m=b+Bf$, let $\widehat m^{C}=L^\dagger h+P_C\widehat m$, which is again a formulation, with $b^C=L^\dagger h+P_Cb$ and $B^C=P_CB$. Then for every $f$,
\[
  \widehat m^{C}-m^*=P_C(\widehat m-m^*),\qquad
  \|\widehat m-m^*\|^2-\|\widehat m^{C}-m^*\|^2=\|\Pi_R(\widehat m-m^*)\|^2 .
\]
If the constraint holds only approximately, the gain becomes $\|\Pi_R(\widehat m-m^*)\|^2-\|L^\dagger h-\Pi_Rm^*\|^2$, which can be negative.
\end{proposition}

The proof writes $m^*=L^\dagger h+P_Cm^*$ and uses the pointwise orthogonality of $\Pi_R$ and $P_C$. Because the identity holds for every $f$, imposing the constraint at inference on a network trained without it also cannot increase the field error on the support of $p_t$; sampling that leaves it is the subject of \cref{sec:decoded}.

\subsection{Measuring the burden}\label{sec:measure-burden}

To say how much of a target is a burden, we need a learner with limited capacity. We use the class $\Fc_d$ of networks whose per-token output is $Wh_i(z,t)+\mu_0$, where $W\in\R^{n\times d}$ and $\mu_0$ are shared across tokens and times and $h_i$ is an arbitrary function of the whole input. A diffusion transformer with patch size one and no bypass from input to output belongs to $\Fc_d$, with $d$ its width. Because $W$ is shared, every output lies, up to the bias, in a common subspace of dimension at most $d$. With time weights $q_t$, the \emph{reducible error} of a formulation is
\[
  \Ec_d(b,B)=\inf_{f\in\Fc_d}\ \E_{t,i}\bigl[q_t\,\|b+Bf-m^*\|^2\bigr],
\]
and $\tail_d(M)=\sum_{j>d}\lambda_j(M)$ denotes the sum of the eigenvalues of $M$ beyond the $d$-th. When the network's target is a single point and $B=\gamma_tI$, its weighted variance outside the best $d$-dimensional subspace is error that no choice of $h_i$ removes: for finite total weight, the infimum equals $\tail_d$ of the weighted second moment of that target about its weighted mean, by the Eckart--Young theorem in reducible-error form (\cref{lem:eckart}; \citet{rae} state the raw-loss version). When the target is a fiber, as for a projection, the tail computed from any one optimal output bounds it from above; a direction-dependent $B$ makes it a weighted low-rank problem. At $d\ge n$, $\Ec_d$ vanishes for every Bayes-consistent formulation, since a head of rank $n$ realises any output.

\section{Relation I: one burden, many fixes}\label{sec:relation-i}

The fixes in \cref{tab:formulations} were proposed for different reasons: geometry, manifold structure, width, bandwidth. As far as the pass-through burden is concerned, they act on the same object.

\begin{theorem}[One burden, many fixes]\label{thm:relation-i}%
\leavevmode
\begin{enumerate}
\item \emph{The pass-through term is radial.} For any linear path and any data, $\kappa_tz$ is parallel to $z$, so $\zh\zh^\top m^*=\kappa_tz+\beta_t\,\zh\zh^\top D$, whether or not the representation lives on a sphere. If both endpoint norms are fixed, $|x|=|\varepsilon|=R$, then on the chord path the whole radial component is determined by the known function $z^\top(\varepsilon-x)=h_m$ of \eqref{eq:sphere-h}; %
if only one endpoint norm is fixed, for instance with Gaussian noise, it is not determined by $z$.
\item \emph{Clean-data prediction and tangent projection remove the same term.} With $\widehat v=(z-\hat x)/t$, the Bayes-optimal output is $\hat x=D$, and the pass-through term $z/t$ is supplied entirely by the conversion. On the geodesic path with time $s$, let $\Omega$ be the angle between the endpoints and consider the loss $\psi\,\|P_T(z)f-u^S\|^2$ with a weight $\psi(\Omega,s)>0$; $\psi\equiv1$ is unweighted Riemannian flow matching. The loss is minimised by the field $-P_T(z)\,\E_\psi[c\,x\mid z]$, where $c=\Omega/\sin(s\Omega)$ and $\E_\psi$ is the posterior reweighted by $\psi$; for $\psi\equiv1$ this is the Bayes field $m^S$ of the geodesic path. The optimal outputs form $\{-\E_\psi[c\,x\mid z]+\lambda(z,s)\,\zh\}$, a set that contains $g^*_\psi=-\E_\psi[c\,x\mid z]$, which has no pass-through term; the pass-through term and every radial component are supplied by the projection.
\item \emph{The norm constraint makes the removal lossless.} Let $B=\beta_tP_T$. Without radial compensation ($b=0$) the formulation leaves the error $\E\|\zh\zh^\top m^*\|^2$, which is generally nonzero. On the chord path the compensation $b=L^\dagger h_m$ makes the formulation Bayes-consistent exactly when $Lm^*=h_m$, which the exact norm constraint guarantees. On the geodesic path the radial target is zero, so $b=0$ suffices.
\end{enumerate}
\end{theorem}

\begin{figure}[t]
\centering
\includegraphics[width=\linewidth]{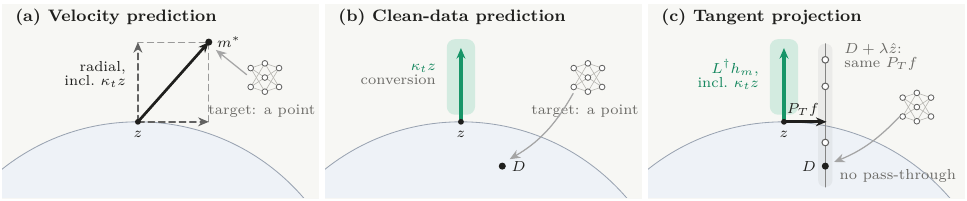}
\caption{\textbf{One burden, many fixes.} Schematic of \cref{thm:relation-i} at a point $z$ on the chord path, drawn for data on the sphere. Panels (a) and (b) hold for any data; the norm constraint enters only in (c), where it makes discarding the radial output lossless. Ink marks what the network must output (the icon points at its optimal output); green marks what the analytic operations supply. (a) Velocity prediction: the network's optimal output is $m^*=\E[U\mid z]$, including the radial part, which contains the pass-through term $\kappa_tz$. (b) Clean-data prediction: the network outputs $D$, and the conversion supplies $\kappa_tz$. (c) Tangent projection: only $P_Tf$ enters the field, so every output on the shaded segment, the fiber $\{D+\lambda\zh\}$ of optimal outputs, gives the same field; the network can take $D$, a target without the pass-through term, and the compensation $L^\dagger h_m$ supplies the radial component, including $\kappa_tz$.}
\label{fig:relation-i}
\end{figure}

The geodesic path and the projection act at different places. The path changes the required dynamics itself: its target is tangent, and its pass-through coefficient is $\Omega\cot(s\Omega)$, near zero at the noise end, where $\Omega$ concentrates near $\pi/2$ (\cref{prop:chord-slerp} relates the two paths exactly). The projection changes the network's task: its fiber contains an output without the pass-through term (\cref{fig:relation-i}). Proofs are in \cref{app:proofs}; (b) covers the weighting $\psi=\mathrm{sinc}^2((1-s)\Omega)$ used by \citet{rjf}.

The norm constraint therefore does not create the burden; it is what makes discarding the radial output lossless. This refines the geometric readings of \citet{rjf}, who attribute the failure of Euclidean flow matching to geometric interference, and of \citet{rit}, who call it a pathology specific to DINOv2. Both place the difficulty in the radial component, which contains the pass-through term for any data; their two remedies remove that same term.

\subsection{What the pass-through term costs a finite network}

The pass-through term is an algebraic component of the target, and the data term can cancel it: for standard Gaussian data and noise at $t=1/2$ on the chord path, $D=z$ and the velocity field $m^*=\kappa_tz+\beta_tD$ vanishes. What the term costs a finite network therefore depends on the representation. In directions the data do not occupy nothing cancels it.

\begin{proposition}[Pass-through floor]\label{prop:pt-floor}%
For a linear path with $\Cov(\varepsilon)=I$,
\[
  \Cov_t(m^*)=\kappa_t^2\sigma_t^2I+(\kappa_t\alpha_t+\beta_t)^2\Sigma_x-\beta_t^2P_t,\qquad P_t=\E\,\Cov(x\mid z),
\]
and on the null space of $\Sigma_x$ the field covariance equals $\kappa_t^2\sigma_t^2$. For velocity prediction on the chord path with $t$ uniform and unit weights, every direction the data do not occupy costs $1$, and for any data
\[
  \Ec_d\ \ge\ \Bigl(1-\frac{\pi^2}{16}\Bigr)(n-d)\ \approx\ 0.383\,(n-d).
\]
Clean-data prediction has no data-independent lower bound of this kind.
\end{proposition}

The proof combines \cref{lem:eckart} with $P_t\preceq\Sigma_x-\Cov(\Dlin)$, the fact that the MMSE does not exceed the linear MMSE, with equality when $D$ is affine. For a direction of variance $\lambda$ the per-direction cost averaged over $t$ is $1+\lambda-\frac{\pi}{2}\sqrt\lambda$, whose minimum, at $\lambda=\pi^2/16$, is the constant. The per-direction expression coincides with the affine velocity coefficients of \citet[Appendix B.4]{rsb}.

In a single-image width sweep, \citet[Figure~4]{rjf} find the angular loss levelling off well above the radial loss at the two narrower widths, with both vanishing at widths $768$ and $896$, and attribute this to the Euclidean objective forcing the network to fit the radial component. The rank limit already produces this pattern. For a single image the irreducible error vanishes, and the rank-optimal residual of velocity prediction is the noise in the $n-d$ directions the head does not span \citep[Section~4.1]{rae}; although the pass-through term $z/t$ is radial, this residual is mostly angular (\cref{app:proofs}). A stalled angular loss therefore need not mean that fitting the radial component takes capacity from the angular one.

Clean-data prediction avoids this rank cost but pays elsewhere: its velocity error is $\beta_t$ times its clean-data error, so velocity-loss training of clean-data prediction weights the clean-data loss by $1/t^2$, and in a common velocity metric%
\begin{equation}\label{eq:gain}
  \Cov_t(m_v)-\beta_t^2\Cov_t(D)=I-\frac{1-t^2}{t^2}\,\Sigma_x :
\end{equation}
the velocity target has the larger field in directions with variance below $t^2/(1-t^2)$, and the clean-data target in the others. Removing the pass-through term thus trades a rank cost for an error gain that grows towards the data end, and which formulation has the lower bound depends on the spectrum and the time weights.

\subsection{Reading the literature}\label{sec:reading-i}

Two further instances are read in \cref{app:cases}: the endpoint-induced spherical velocity of \citet{speech} has the structure of \cref{thm:relation-i}(b), and the radial share of the velocity that \citet{aligning} measure in VAE latents is close to the energy share of the pass-through term at the noise end. Sources are in \cref{tab:case-table}.

\paragraph{Bandwidth and learned skips.} At width equal to the token dimension, \citet{rsb} find that a noise-dependent target makes the residual stream carry the input to the readout. Widening the stream at fixed depth and number of tokens nearly closes the gap between velocity and clean-data prediction on DINOv2-B latents (\cref{tab:beyond}), %
and, in their pixel-space experiments, a learned scalar skip, written in our notation as $\widehat m=\kappa(t)z+\gamma(t)f$ and initialised at $\kappa=0$, learns $\kappa(t)\approx1/t$, the analytic pass-through coefficient $\kappa_t$, and improves FID from $139.83$ to $9.99$: the network hands the pass-through term to an analytic operation by itself. %
At $d=n$ the pass-through term is therefore still the burden, paid for in bandwidth rather than output rank.

\section{Relation II: same information, different tasks}\label{sec:relation-ii}

\Cref{sec:relation-i} showed that operations built on different information can remove the same burden. The converse also holds: the same information, implemented by different operations, can leave the network different tasks. We show this on the sphere, where the norm constraint supplies the radial dynamics exactly.

\paragraph{Setting.} To compare three formulations, we use the chord path with uniform spherical noise and data on the sphere, $|x|=|\varepsilon|=R=\sqrt n$ \eqref{eq:sphere-h}, with the velocity target, so $\kappa_t=1/t$ and $\beta_t=-1/t$.
\begin{itemize}
\item[(E)] \emph{Exact compensation}: $\widehat m_{\mathrm E}=L^\dagger h_m+\beta_tf$, the radial part supplied exactly by the norm constraint and the network output added to it.
\item[(P)] \emph{Projection}: $\widehat m_{\mathrm P}=L^\dagger h_m+\beta_tP_Tf$, the same compensation with the output projected onto the tangent space; it is (E), and equally clean-data prediction, with the norm constraint imposed (\cref{prop:impose}); note that $P_Tz=0$.
\item[(S)] \emph{Scalar offload}: $\widehat m_{\mathrm S}=\kappa_tz+\beta_t(\cbar z+f)$, that is $b_{\mathrm S}=(\kappa_t+\beta_t\cbar)z$ and $B_{\mathrm S}=\beta_tI$, with $\cbar=(1-t)/\nu_t$ and $\nu_t=(1-t)^2+t^2$. This is the LMMSE coefficient computed from the second moment $\E|x|^2=R^2$ alone, an EDM-type skip connection.
\end{itemize}

\begin{theorem}[Same information, different tasks]\label{thm:relation-ii}%
\leavevmode
\begin{enumerate}
\item \emph{The tasks differ.} Under (E) the network's target is the single point $f_{\mathrm E}^*=P_TD$. Under (P) it is the fiber $\{f:\ P_Tf=P_TD\}$, which contains $D$. If $x$ lies in an $r$-dimensional subspace $S$, then $D\in S$, while $P_{S^\perp}f_{\mathrm E}^*=-c_{\rm ex}\,t\,P_{S^\perp}\varepsilon$ with $c_{\rm ex}=h_D/|z|^2$; this component has zero mean and covariance $t^2e_{\rm ex}(t)\,I_{n-r}$ with $e_{\rm ex}(t)>0$ for $0<t<1$.
\item \emph{The costs differ.} For any data on the sphere, any $d$ and any time weights, $\Ec_d(\mathrm P)\le\Ec_d(\mathrm E)$. Let $x$ lie in an $r$-dimensional subspace $S$, with a mean that need not vanish, and set $e_{\mathrm E}=\E_t[q_t\beta_t^2t^2e_{\rm ex}(t)]$. Then
\[
  \Ec_d(\mathrm E)\ \ge\ (n-d-r)_+\,e_{\mathrm E},\qquad
  \Ec_d(\mathrm P)=\Ec_d(\text{clean-data prediction})=0\quad\text{for } d\ge r .
\]
In particular, for $r\le d<n-r$ and time weights positive on a set of positive measure, $\Ec_d(\mathrm P)<\Ec_d(\mathrm E)$. %
\item \emph{The scalar offload: almost the same information and cost.} At $t=1/2$, $\cbar=1$ and $h_D=|z|^2$, so the radial part supplied by the norm constraint, $L^\dagger h_D=z$, equals that of the scalar statistic, and (S) coincides with (E): in denoiser units $\widehat D_{\mathrm S}=\widehat D_{\mathrm E}=z+f$, whereas $\widehat D_{\mathrm P}=z+P_Tf$. For $t\ne1/2$ and $n>3$, let $\eta^{\rm rad}_t$ be the energy of the radial field error of the scalar offload as a fraction of the energy of the exact radial component $L^\dagger h_m$. Then $\eta^{\rm rad}_t\le4/(n+1)$, a bound approached as $t\to1/2$. Under (S) the network's target is the single point $f_{\mathrm S}^*=D-\cbar z$. If $x$ lies in such a subspace, then $P_{S^\perp}f_{\mathrm S}^*=-\cbar\,t\,P_{S^\perp}\varepsilon$ with $\Cov(P_{S^\perp}f_{\mathrm S}^*)=\cbar^{\,2}t^2\,I_{n-r}$, and with $e_{\mathrm S}=\E_t[q_t\beta_t^2\cbar^{\,2}t^2]$
\[
  \Ec_d(\mathrm S)\ \ge\ (n-d-r)_+\,e_{\mathrm S} .
\]
Under velocity-loss weighting ($q_t=1$, $t$ uniform), $e_{\mathrm S}=\E_t[\cbar^{\,2}]=\pi/4$ and $\frac{\pi}{4}\le e_{\mathrm E}\le\frac{\pi}{4}\frac{n}{n-1}$.
\end{enumerate}
\end{theorem}

Formulations (E) and (P) isolate the effect of the output map. Under (E) the network must itself remove the radial part $(\zh^\top D)\zh$ of $D$; since $\zh$ ranges over all directions, this correction is proportional to $z$ and puts variance into every direction the data do not occupy. Under (P) the radial output is discarded. Clean-data prediction, which uses neither the second moment nor the constraint, also avoids this cost, and imposing the constraint on it gives (P), so $\Ec_d(\mathrm P)\le\Ec_d(\text{clean-data prediction})$ for every $d$ (\cref{prop:impose}). The radial information is therefore neither necessary for avoiding the cost nor sufficient, since (E) uses it exactly and still pays: what decides is how it is implemented.

The difference needs a projection that varies with $z$. For a constraint with fixed $L$, projecting the output changes nothing.

\begin{proposition}[Affine constraints]\label{prop:affine-constraints}%
Let $Lx=c$ almost surely with fixed $L$ and $c$, so that $h=\kappa_tLz+\beta_tc$ and $P_C=I-L^\dagger L$ is a fixed matrix.
\begin{enumerate}
\item Every distribution with the same mean and covariance satisfies the constraint, and $La^*=h$.
\item If $\Pi_Rb=L^\dagger h$, then $\Ec_d(b,\gamma_tP_C)=\Ec_d(b,\gamma_tI)$ for every $d$ and all time weights: projecting the network output onto the unconstrained directions leaves the same reducible error as adding it.
\end{enumerate}
\end{proposition}

A fixed projection maps $\Fc_d$ into itself, so it can only restrict what the network outputs; this is why (b) holds, and by (a) it applies to the full LMMSE offload $b=a^*$, on which imposing an affine constraint leaves the reducible error unchanged. The tangent projection $P_T$ varies with $z$, so $P_T(z)f$ can lie outside $\Fc_d$; this is what lets (P) leave the network a smaller task than (E).

\begin{figure}[t]
\centering
\includegraphics[width=\linewidth]{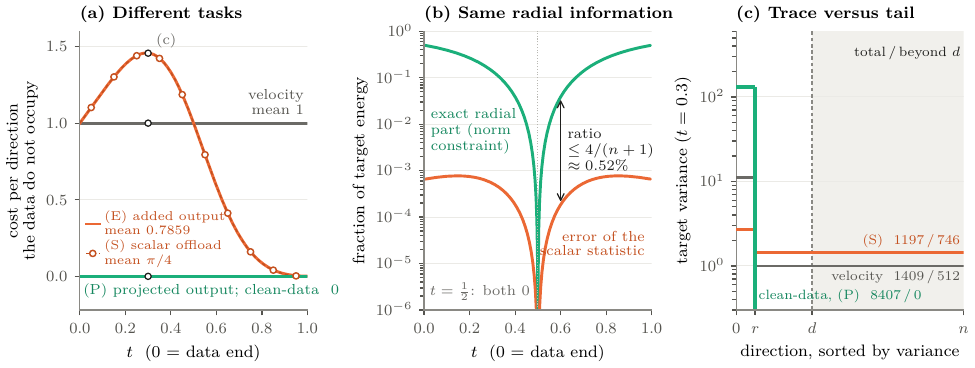}
\caption{\textbf{Same information, different tasks.} Relation II on the sphere (chord path, spherical noise, $n=768$, velocity target). (a) Cost per direction the data do not occupy, the target variance in field units, for (E) and (P), which share the exact radial compensation, and for (S) (dashed, open markers), whose time averages under velocity-loss weighting are $e_{\mathrm E}$ and $e_{\mathrm S}$ (\cref{thm:relation-ii}(b), (c)); the markers at $t=0.3$ are the levels of panel (c) beyond $r$. (b) Energies of the exact radial component $L^\dagger h_m$, which the norm constraint supplies, and of the radial error of the scalar statistic, as fractions of $\E|U|^2=2R^2$ (\cref{thm:relation-ii}(c)). (c) At $t=0.3$, for data uniform on a sphere in an $r=64$-dimensional subspace: sorted eigenvalues of the second moment of the network's target in field units, for velocity prediction ($c=1$), the scalar offload (S) ($c=\cbar$) and clean-data prediction ($c=0$) in the family $\widehat D=cz+c_{\rm out}f$, and for the optimal output $D$ of (P). (S) has the smallest total, the quantity \eqref{eq:trace} minimises, and the largest part beyond $d=256$. Closed forms, except $\E|D|^2$ in (c) (Monte Carlo).}
\label{fig:relation-ii}
\end{figure}

The scalar offload carries almost the exact radial information (\cref{fig:relation-ii}b), yet in the directions the data do not occupy it costs almost as much as exact compensation (\cref{fig:relation-ii}a). Its saving over velocity prediction is not uniform in time: near the data end it leaves more in these directions than velocity prediction, so the net effect depends on the time weighting.

\subsection{Trace versus tail}

Standard preconditioning chooses its skip coefficient by the total variance of the network's target. Over the scalar affine family $\widehat D=cz+c_{\rm out}f$ in denoiser units, orthogonality gives $\E\|x-cz\|^2=\E\|x-D\|^2+\E\|D-cz\|^2$, so for centred data the total variance of the network's Bayes target, scaled by $c_{\rm out}^2$, is minimised by the scalar LMMSE coefficient%
\begin{equation}\label{eq:trace}
  c=\frac{(1-t)\bar\lambda}{(1-t)^2\bar\lambda+t^2},\qquad \bar\lambda=\tr\Sigma_x/n ,
\end{equation}
which coincides with the $c_{\rm skip}$ that EDM chooses to minimise the amplification of network errors, after the change of variables $\sigma=t/(1-t)$, $\sigma_{\rm data}^2=\bar\lambda$ \citep[Appendix B.6]{edm}. The trace criterion is optimal under EDM's error model, in which network errors are isotropic with a fixed relative scale.

For a network with an output-rank limit, what matters is the tail of the target's spectrum. In the affine family, to which (S), velocity prediction and clean-data prediction belong, the second moment of the network's target splits into a part the formulation controls and a part it cannot touch.

\begin{proposition}[Reducible error in the affine family]\label{prop:affine-tail}%
Let $b$ be affine and $B=\gamma_tI$ with $\gamma_t\ne0$ wherever $q_t>0$, and assume $0<\E_t[q_t\gamma_t^2]<\infty$. Define
\[
  \Rbar=\E_{t,i}\bigl[q_t\,r^*r^{*\top}\bigr],\qquad
  \Abar(b)=\E_{t,i}\bigl[q_t\,(a^*-b-\gamma_t\mu)(a^*-b-\gamma_t\mu)^\top\bigr],
\]
where $\mu$ is the $q_t\gamma_t^2$-weighted mean of the network target, and assume both are finite. Then:
\begin{enumerate}
\item $\Ec_d(b,\gamma_tI)=\tail_d\bigl(\Abar(b)+\Rbar\bigr)$;
\item at a fixed path and time weights $q_t$, $\Rbar$ is the same for every member of the family, and $\Abar(b)$ depends only on the first two moments of $Q$ and on $(b,\gamma_t)$;
\item the minimum over the family is $\tail_d(\Rbar)$, attained by the full LMMSE offload $b=a^*$;
\item if $d\ge n$, every tail vanishes, so $\Ec_d=0$.
\end{enumerate}
\end{proposition}

Because $r^*$ is orthogonal to every affine function of $z$, the formulation controls only $\Abar(b)$, and the full LMMSE offload is optimal in the family for any data, also against members with infinite total weight such as (S) under the velocity loss (\cref{app:proofs}); \citet{hybridspectral} derive this optimality for their Wiener mixture under Gaussian data. On steep spectra the trace and the tail disagree (\cref{thm:relation-ii}(c), \cref{fig:relation-ii}c): in field units, the trace-optimal scalar leaves variance $c^2$ in every direction of zero data variance. The full LMMSE offload has reducible error $\tail_d(\Rbar)$, the minimum over the family (\cref{prop:affine-tail}(c)).

\paragraph{General spectra.} Beyond low-rank data, \cref{prop:general-spectrum} gives a sufficient condition at a fixed time under which the projection has a smaller reducible error than the scalar offload. The condition compares a lower bound on the scalar offload's reducible error with an upper bound on the projection's; per direction, the term of the first exceeds that of the second for variances below $\lambda^*(t)$, about $0.155$ at $t=0.5$ (\cref{fig:general-spectrum}); near the data end the condition fails, because clean-data-type targets have the larger field there \eqref{eq:gain}. On the representation latents of \cref{app:latents}, the condition holds only for the MAE encoder outputs without standardisation, at intermediate $t$, and does not decide the comparison in the other five cases. %

\paragraph{Other settings.} On the geodesic path, the projected family still attains the minimiser of its training loss when $d$ is at least the rank of the data, with or without the weighting of \citet{rjf} (\cref{prop:projection-bound}); the trace criterion \eqref{eq:trace} and the lower bound for (S) in \cref{thm:relation-ii}(c) do not require the sphere. %

\paragraph{Evidence from the literature.} On $16\times16$ pixel patches with width equal to the patch dimension, \citet{jit} report that an EDM-style preconditioner, whose skip coefficient is exactly the scalar LMMSE with $\sigma_{\rm data}=0.5$, reaches FID $35.49$, between clean-data prediction ($8.62$) and velocity prediction ($96.53$), all under the velocity loss. %
At this width \citet{rsb} measure the burden by the spectrum of the network's target, and in the directions the data barely occupy the scalar offload keeps part of the pass-through term (\cref{fig:relation-ii}a), consistent with a result between the two.

\section{Relation III: when statistics and geometry overlap or complement}\label{sec:relation-iii}

The two kinds of known information in \cref{def:known} constrain the required dynamics differently. Using more of one need not bring a formulation closer to the other.

\begin{figure}[t]%
\centering
\includegraphics[width=\linewidth]{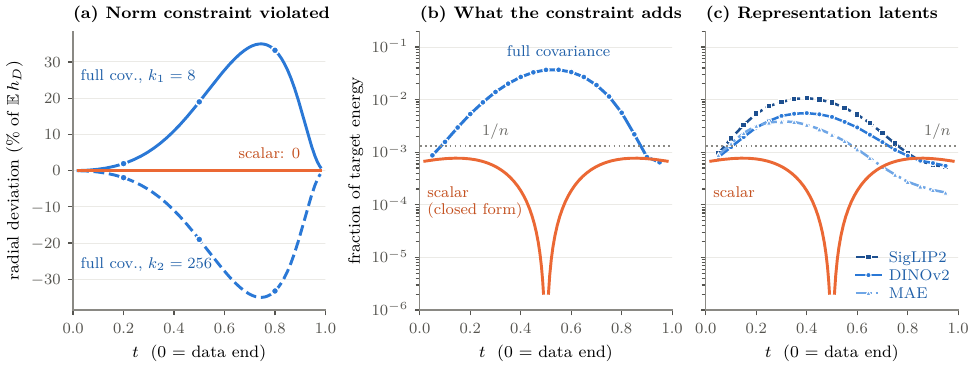}
\caption{\textbf{Overlap or complement.} Relation III on the constructed two-cluster data of \cref{ex:two-clusters} (a, b) and on representation latents (c); spherical noise, chord path. Only the complement is shown: for an affine constraint and the full covariance, the quantity in (b) and (c) is zero (\cref{prop:affine-constraints}(a)). (a) Cluster-conditional radial deviation of the full-covariance LMMSE from the norm constraint, relative to $\E h_D$ (closed form). (b) The part of the dynamics that the constraint determines beyond the statistics, $\E\|L^\dagger(h-La^*)\|^2$ with $a^*$ the prediction from the respective statistics (\cref{thm:relation-iii}); it equals the radial field error of the statistics-only offload. Shown as a fraction of the target energy of the velocity target: full covariance by Monte Carlo ($2\times10^4$ samples per $t$), scalar statistic in closed form; the dotted line is the average energy per direction, $1/n$. (c) The same quantity on the DINOv2-B, SigLIP2-B and MAE-B latents of representation autoencoders, without standardisation, with $a^*$ the LMMSE prediction from their mean and covariance (Monte Carlo; \cref{app:latents}); the scalar curve is that of (b).} %
\label{fig:relation-iii}
\end{figure}

\begin{example}[Full covariance can violate the norm constraint]\label{ex:two-clusters}%
Use the chord path and spherical noise of \cref{sec:relation-ii}, and let the data be an equal mixture of two clusters, uniform on the radius-$R$ spheres of two orthogonal subspaces $S_1$ and $S_2$ of dimensions $k_1\ne k_2$. The mean is zero, $\Sigma_x=\lambda_1P_{S_1}+\lambda_2P_{S_2}$ with $\lambda_i=R^2/(2k_i)$, and the norm constraint holds. Conditional on the cluster, the radial component of the full-covariance LMMSE deviates from the constraint by amounts equal in size and opposite in sign for the two clusters, and nonzero when $k_1\ne k_2$ (\cref{app:proofs}): for $n=768$, $k_1=8$ and $k_2=256$, by $\pm19\%$ of the mean of $h_D$ at $t=0.5$ (\cref{fig:relation-iii}a). The scalar statistic of \cref{thm:relation-ii}(c), which uses less information, has zero cluster-conditional deviation at every $t$, and its radial error energy stays below the average energy per direction, which that of the full covariance exceeds (\cref{fig:relation-iii}b).
\end{example}

To say when a constraint determines something the statistics do not, combine the analytic parts that each kind of information determines into the \emph{analytically available component} $k_0:=L^\dagger h+P_Ca^*$, which can always be written from $\Kc$; with only statistics $k_0=a^*$, and with only a constraint $k_0=L^\dagger h$. %

\begin{theorem}[Overlap or complement]\label{thm:relation-iii}%
Let $r^*=m^*-a^*$. Then $\Pi_R r^*=L^\dagger(h-La^*)$. If $La^*=h$ almost surely, the constraint determines no component of $m^*$ beyond the affine prediction, and the two overlap; otherwise it additionally determines $L^\dagger(h-La^*)$, and the two complement each other. Moreover $m^*-k_0=P_Cr^*$ and
\[
  \|r^*\|^2-\|P_Cr^*\|^2=\|L^\dagger(h-La^*)\|^2 ,
\]
so using both kinds of information leaves exactly this much less to the data than using statistics alone.
\end{theorem}

Proofs are in \cref{app:proofs}. Overlap is a property of the field of a given distribution, not of the known facts alone: second-order statistics imply every affine constraint (\cref{prop:affine-constraints}(a)) but not a norm constraint, since an equal mixture of the origin and the sphere of radius $\sqrt2R$ has the mean and covariance of isotropic data on the sphere and, at $t=1/2$, the radial velocity $z^\top v=\pm R^2/2$ instead of zero. Overlap is also distinct from computational redundancy: at $t=1/2$ the scalar statistic already gives the exact radial velocity, zero, for any data on the sphere, yet on low-rank data the projection has the smaller reducible error (\cref{thm:relation-ii}(b), (c)). Relation~III concerns which components the known information determines, Relation~II how the way it is used changes the network's task.%

\paragraph{Subspace: overlap.} For the subspace, as for every affine constraint, $L$ is fixed and $La^*=h$ (\cref{prop:affine-constraints}(a)): the information overlaps, as also exploited by \citet{pi}, who recover the subspace from the data second moment. %
The structure behind the subspace analysis of \citet{diffusibility} is thus available to an LMMSE offload.

\paragraph{Sphere: complement.} On the sphere $L=z^\top$ varies with $z$, and the norm constraint complements the statistics whenever $La^*\ne h$ with positive probability, as in \cref{ex:two-clusters}. Combining the two analytic parts, as $k_0$ does, then need not extract everything that the two kinds of information determine together; on the sphere, what they determine together generally depends on moments of $z$ beyond the second (\cref{prop:composition,prop:commute,prop:noncommute}). %

\paragraph{Representation latents.} We observe the same complement on the tokens of the three frozen encoders used by representation autoencoders, \mbox{DINOv2-B}, \mbox{SigLIP2-B} and \mbox{MAE-B}, whose tokens lie on the sphere of radius $\sqrt n$ (\cref{sec:known-information}; $2{,}000$ ImageNet images, \cref{app:latents}). With $a^*$ the LMMSE prediction from the mean and covariance, the part of the dynamics that the constraint determines beyond these statistics exceeds the average energy per direction at intermediate noise levels and peaks at $4.3/n$ for DINOv2, $8.2/n$ for SigLIP2 and $3.0/n$ for MAE, whereas the scalar second moment stays below $0.6/n$ at every $t$ (\cref{fig:relation-iii}c, \cref{tab:latents}). %
The LMMSE has the smaller total error, yet its radial component agrees worse with the constraint, on real latents as in \cref{ex:two-clusters}.

\paragraph{Implication.} In \cref{ex:two-clusters}, and to a smaller degree on these latents, the constraint supplies the exact radial component and the anisotropic statistics the tangential structure. The formulation $b=k_0$ with $B=\beta_tP_C$ uses both and is Bayes-consistent (\cref{thm:relation-iii}); it leaves the network only the residual $P_Cr^*$.

\section{From learning difficulty to decoded consequence}\label{sec:decoded}

Sections~\ref{sec:relation-i}--\ref{sec:relation-iii} concern the learning side: which part of the required dynamics a network must implement, and at what cost. The generated latent is then decoded into an image, and image quality is what is evaluated. Let a sampler integrate $\dot z=\widehat m(z,t)$ from $t=1$ to $t=0$, let $z^*_t$ be the trajectory of $m^*$ from the same initial noise, and let $\Phi(s,t)$ be the Jacobian of the flow map of $m^*$ from time $t$ to time $s$. For a small field error $e_t=\widehat m-m^*$, with the flow map and the decoder differentiable along the trajectory and discretisation ignored, the first-order decoded consequence is%
\begin{equation}\label{eq:decoded}
  \delta z_0\approx\int_1^0\Phi(0,t)\,e_t(z^*_t)\,dt,\qquad \delta I\approx J_G(z^*_0)\,\delta z_0 ,
\end{equation}
where $J_G$ is the Jacobian of the decoder. The learning difficulty of a component, measured as in \cref{sec:measure-burden}, involves neither $\Phi$ nor $J_G$. Two field errors of the same size can therefore have different decoded consequences, and a component that costs the network nothing to learn can still matter for the image. We first measure how the decoder factor $J_G$ weights errors by direction; three situations then follow from the preceding sections, and \cref{tab:decoded} in \cref{app:cases} lists published instances. %

\begin{figure}[t]
\centering
\includegraphics[width=\linewidth]{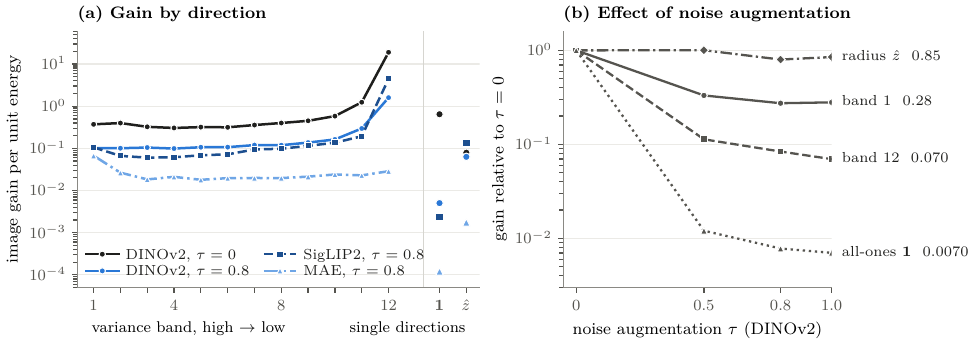}
\caption{\textbf{Decoded consequence depends on direction.} Local gain of representation-autoencoder decoders at data latents: mean squared change of the decoded image (pixel values in $[0,1]$) per unit mean squared latent perturbation, central difference at $1\%$ of the token scale, averaged over $250$ ImageNet images. (a) Twelve bands of principal directions of the token covariance, $64$ per band and $63$ in the last (without the all-ones direction), from highest to lowest variance (perturbations tangent to each token), and two single directions: the all-ones direction $\mathbf 1$, which the encoder output never populates, and the radius $\hat z$. Decoders trained with noise augmentation of strength $\tau$ \citep{rae}; $\tau=0$ is clean training. (b) DINOv2-B decoders: gain relative to $\tau=0$; labels give the values at $\tau=1.0$. Ratios with $95\%$ intervals are in \cref{tab:decoder}.}
\label{fig:decoder}
\end{figure}

\paragraph{The decoder factor, measured.}%
The formulations of \cref{sec:relation-ii} differ in how much error they leave in the directions the data barely occupy, so we compare those directions with the principal ones. We measured the local gain of the released decoders of representation autoencoders at data latents, the mean-squared change of the decoded image per unit mean-squared latent perturbation (\cref{fig:decoder}; protocol in \cref{app:decoder}). For the DINOv2-B decoder trained only on clean latents, a perturbation in the lowest-variance band produces $51$ times the mean-squared image change of one of the same energy in the highest-variance band (\cref{tab:decoder}). %
Noise-augmented decoder training lowers the gain in every measured direction, most in relative terms along the all-ones direction of \cref{sec:known-information} (\cref{fig:decoder}b), and the low-variance directions keep $13$ to $17$ times the gain of the principal ones. %
The SigLIP2-B decoder behaves like the DINOv2-B one, with a factor of $42$; the MAE-B decoder does not, with $0.43$. %

\paragraph{Known components.}%
Under the projection with an exact norm constraint, with radial compensation on the chord path or on the geodesic path, the radial field error is zero on the support of $p_t$ for every network (\cref{prop:impose}), so the radial component carries no learning burden. Its decoded consequence need not be zero. The terminal radius is fixed by the sampler and by any normalisation applied at inference, and a radial deviation $\delta r$ moves the image by $J_G\zh\,\delta r$. Locally, the six decoders of \cref{tab:decoder} have a gain for radial perturbations $3$ to $25$ times smaller than for tangential ones of the same energy; \cref{tab:decoded} lists the same finding for VAE decoders. %
Large radial changes still matter: the $6.77$ quoted in \cref{sec:intro} is the best value of a sweep over the radius to which generated latents are projected before decoding, reached at $R\approx45$, whereas the data norm $R\approx27.7$ gives $7.79$ with the same learned field \citep[Section~5.3]{rjf}.
Whether the error in a known, unlearned component is negligible is thus a property of the sampler and the decoder, not of the learning problem.

\paragraph{Directions the formulation forces the network to fill.}%
Under an output-rank limit, formulations whose target is full rank leave error that the network cannot remove in the directions the data do not occupy: velocity prediction (\cref{prop:pt-floor}) and the scalar offload of \cref{sec:relation-ii} (\cref{thm:relation-ii}(c)). Propagated through $\Phi$, these errors can become terminal deviations off the data support, with effect $J_GP_{S^\perp}\delta z_0$; a decoder trained only on clean latents is not constrained there, because the encoder rarely or never produces such latents; noise-augmented decoder training \citep[Section~4.3]{rae} is aimed at this gap. For the DINOv2-B and SigLIP2-B decoders, the directional sensitivity measured above can compound this learning-side cost when sampling errors reach the decoder along low-variance directions. %

\paragraph{Analytic components chosen for their consequence.}%
Some analytic components are not derived from known information at all but chosen for what they do to sampling: the normal contraction $-\alpha(t)\,r(z)$ of \citet{msfm} keeps samples near the manifold irrespective of how accurately the field is learned, trading accuracy of the field for control of the trajectory (\cref{tab:decoded} lists a second instance). When the objective is decoded quality rather than fitting the latent distribution, the field to implement need not be $m^*$.

\section{Related work}\label{sec:related}

\paragraph{Prediction targets and preconditioning.} Besides clean-data prediction \citep{jit,diffusibility,rit} and EDM preconditioning \citep{edm}, \citet{jinwang} derive the optimal mixing coefficient between clean-data and noise targets for time-shared linear denoisers and show that noise-level-dependent denoisers remove a penalty proportional to the ambient codimension; that penalty is the pass-through term outside the data subspace. \Citet{hybridspectral} mix targets per direction of the data spectrum, with a Wiener mixture that is optimal for Gaussian data. \Citet{asymflow} keep the pass-through term only in a principal subspace of each patch, and \citet{anisoflow} weight it per direction with learned weights; both are formulations with a direction-dependent $B$, beyond the family $B=\gamma_tI$ of \cref{prop:affine-tail}.

\paragraph{Geometric and manifold-constrained generation.} Riemannian flow matching \citep{riemannianfm} provides geodesic paths and tangent-space velocities. \Citet{rjf} apply it to hyperspherical representation latents, \citet{aligning} to spherical variational latents, and \citet{speech} to spherical speech tokens; \citet{msfm} add an analytic contraction towards a data manifold.

\paragraph{Architecture and bandwidth.} \Citet{rae} derive a lower bound on network width for representation latents and introduce a wide head. \Citet{jlt} draw the distinction of \cref{prop:fiber} for clean-latent versus velocity prediction on VAE latents and compare the two under a matched velocity objective. \Citet{rsb} show that changing the target, changing the architecture and widening the residual stream act on one burden, the input that the stream carries to the readout (\cref{sec:reading-i}), and describe clean-data prediction as an analytic skip.

\paragraph{Measuring latent difficulty.} Diffusability measures based on the irreducible error of the Bayes-optimal field \citep{latentdiffusability} or on the MMSE along the path \citep{fishergeometry} quantify what cannot be predicted at all, whereas our reducible-error split (\cref{prop:affine-tail}) concerns what a finite network can, but may fail to, represent. Representation-alignment objectives for tokenizers \citep{vavae} aim to make latents easier to generate without separating these parts.

\paragraph{Learning dynamics of denoisers.} Trained denoisers approach their linear (Gaussian) part in the generalisation regime \citep{hiddengaussian}, and the learning dynamics of linear denoisers show a spectral bias \citep{spectralbias}. These results bear on the size of the residual $\Rbar$ and on the optimisation side of difficulty (\cref{sec:disc-beyond}).

\section{Discussion}\label{sec:discussion}

\subsection{Principles for design}\label{sec:disc-design}

\Cref{sec:relation-i,sec:relation-ii,sec:relation-iii,sec:decoded} yield three principles for constructing generative systems.
\begin{enumerate}
\item \emph{Capacity-aware parameterisation.} Choose an analytic offload by the target it leaves relative to the network's capacity, not by how much of the target's variance it removes or how much information it uses (\cref{thm:relation-ii}, \cref{eq:trace}).
\item \emph{Statistical--geometric hybrid formulations.} Combine statistical offloads with exact constraints, which takes off the network whatever the constraint determines beyond the affine prediction, without increasing the field error on the support (\cref{prop:impose}, \cref{thm:relation-iii}). %
\item \emph{Decoder-aware allocation.} Divide the dynamics between analytic operations and the network with the sampler and the decoder in view (\cref{sec:decoded}): where the decoder amplifies the directions the data barely occupy, as the DINOv2-B and SigLIP2-B decoders do (\cref{fig:decoder}), keep those directions out of the network's target. %
\end{enumerate}
Relation~I adds a rule for analysis rather than design: for a new representation or network, compare the width with the token dimension and with the pass-through burden of the chosen target (\cref{prop:pt-floor}) before attributing a failure to the geometry of the latent space.

\subsection{Comparing representations}\label{sec:disc-representations}

At the same output rank and loss weights, changing the formulation can reverse which of two representations has the smaller reducible error. %
For the affine family with finite total weight and a representation with grid distribution $\mathbf Q$ and token marginal $Q$, $\Ec_d(b;\mathbf Q)=\tail_d(\Abar_b(Q)+\Rbar(\mathbf Q))$, where only $\Abar_b$ depends on the formulation (\cref{prop:affine-tail}). %
Two representations can therefore be ranked one way under a formulation and the other way under the full LMMSE offload; \cref{prop:reversal} gives a sufficient condition for such a reversal, and \cref{app:further} a pair of representations for which the reversal can be verified directly.

Comparing representations under a single formulation thus compares a mixture of two things: the second-order structure that the formulation leaves to the network, and the structure beyond per-token affine prediction. Rankings by sample quality also reverse with the formulation beyond the affine family: in the experiments of \citet{rjf}, SigLIP latents are worse than MAE latents under Euclidean flow matching (FID $130.21$ versus $50.48$) and better under their method ($10.39$ versus $19.82$). %
A comparison of representations should therefore state the formulation, or compare errors after a full LMMSE offload, which isolates $\tail_d(\Rbar)$; a comparison of decoded quality depends on the decoder as well (\cref{sec:decoded}).

\subsection{Beyond representability}\label{sec:disc-beyond}

In several reports, the differences between formulations shrink as the width grows relative to the token dimension, and at $d=n$ the evidence of \cref{sec:reading-i} points to residual-stream bandwidth (\cref{tab:beyond} in \cref{app:cases}). Others remain at widths above the token dimension: clean-data prediction still improves on velocity prediction at width $1536$ for tokens of dimension $1024$ and $768$ \citep{diffusibility}, and in the FLUX.2 ablation of \citet{aligning}, with tokens of dimension $128$ and width $768$, the tangent projection still helps; they point to supervision otherwise spent on radial motion, to which the decoder is insensitive. %
The projection makes radial errors free (\cref{prop:fiber}(b)), so all of the network's loss falls on the tangential components.

The analysis points to two directions beyond representability: how bandwidth, loss allocation and the statistical and geometric structure of the latents shape the training of finite networks, and how formulation-dependent field errors reach decoded quality through the sampler and the decoder.

\section{Conclusion}\label{sec:conclusion}

Separating where a learning burden comes from, what the generative system knows about the latents, and which operation uses that knowledge shows that a representation shapes generative learnability twice: through the dynamics it requires, and through the structure that lets parts of those dynamics be realised analytically. Information is not implementation: the same information, implemented by different operations, leaves a finite network different tasks. %
 For representations, the separation means that generative difficulty is not a property of a representation alone: a comparison of representations should state the formulation and the learner's output rank and loss weights, since even when the last two are fixed, changing the formulation can reverse which of two representations has the smaller reducible error. %

\bibliographystyle{tmlr}
\bibliography{references} %

\clearpage %
\appendix
\appendixnumbering

\section{Notation and conventions}\label{app:notation}

\begin{table}[h]
\centering
\small
\begin{tabularx}{\linewidth}{@{}l>{\raggedright\arraybackslash}X@{}}
\toprule
Symbol & Meaning\\
\midrule
$t\in[0,1]$ & time; $t=0$ is the data end (works with data at $t=1$ are converted by $t\mapsto1-t$)\\
$z=\alpha_tx+\sigma_t\varepsilon$ & linear path; chord path $\alpha_t=1-t$, $\sigma_t=t$\\
$x\in\R^n$, $\mathbf Q$, $Q$, $\Sigma_x$, $\lambda_j$ & latent token, law of the clean grid, its token marginal, covariance and eigenvalues of $Q$ (zero mean unless stated)\\
$\varepsilon$ & noise, independent of $x$, $\Cov(\varepsilon)=I$; Gaussian or uniform on the sphere of radius $R=\sqrt n$\\
$U=\kappa_tz+\beta_tx$ & regression target \eqref{eq:target}; $\kappa_tz$ is the pass-through term\\
$m_U$, $m^*$ & Bayes field of a target $U$; required dynamics, the Bayes field of the path velocity $v$\\
$D$, $\Dlin$, $P_t$ & denoiser $\E[x\mid z]$ (conditioned on the noisy grid), LMMSE denoiser, MMSE matrix $\E\Cov(x\mid z)$\\
$a^*$, $r^*$ & affine part $\kappa_tz+\beta_t\Dlin$ and residual $\beta_t(D-\Dlin)$ of $m^*$\\
$L$, $h$, $\Pi_R$, $P_C$ & pointwise relation $Lv=h$; projection onto $\row L$; $P_C=I-\Pi_R$\\
$h_D$, $h_m$ & on the sphere, $z^\top x=h_D$ and $z^\top v=h_m=\kappa_t|z|^2+\beta_th_D$ \eqref{eq:sphere-h}\\
$P_T=I-\zh\zh^\top$ & tangent projection on the sphere\\
$k_\Kc$, $k_0$ & jointly determined (\cref{def:components}) and analytically available (\cref{sec:relation-iii}) components\\
$\widehat m=b+Bf$ & formulation (\cref{def:formulation})\\
$\Fc_d$, $\Ec_d$, $\tail_d$ & output-rank-limited class, reducible error, eigenvalue tail beyond the $d$-th\\
$\Abar(b)$, $\Rbar$, $q_t$ & affine and residual second-moment matrices (\cref{prop:affine-tail}); time weights\\
$\nu_t$, $\cbar$ & $(1-t)^2+t^2$ and $(1-t)/\nu_t$\\
$\Phi(s,t)$, $J_G$ & Jacobian of the flow map of $m^*$ from $t$ to $s$; Jacobian of the decoder\\
\bottomrule
\end{tabularx}
\end{table}

\emph{Grid conditioning.} The path is applied to every token of the grid, turning the clean grid, with law $\mathbf Q$, into a noisy grid $Z$. $Q$ is the law of a token at a uniformly drawn position, and $p_t$ the joint law of the noisy grid $Z$ and such a position $i$, with $z=Z_i$. A network sees the whole noisy grid, so $D$, $m_U$ and $m^*$ condition on $Z$; we write $\E[\,\cdot\mid z]$ to keep notation light. Known statistics are the first two moments of $Q$, and affine functions are affine in the token's own $z$, with coefficients shared across positions; by the tower property, $\E[g(z)D^\top]=\E[g(z)x^\top]$ for every function $g$ of the token's own $z$, so LMMSE quantities are unchanged by grid conditioning. The residual $r^*$ therefore also contains what the other tokens predict linearly about $x$. Pointwise relations hold token by token and therefore also for the grid-conditioned fields. In the constructed examples tokens are independent, and conditioning on $Z$ reduces to conditioning on $z$. All tokens share one network.

\emph{Units and weights.} All statements are in the field metric $\E\|\widehat m-m^*\|^2$, with time weights $q_t$ (\cref{sec:formulations}).

\section{Proofs}\label{app:proofs}

\subsection{Formulations and reducible error}

\begin{lemma}[Reducible error under an output-rank limit]\label{lem:eckart}%
For a target with Bayes field $m_U$ and weights $w_t$ with finite total weight $\E_t[w_t]$, the infimum over $\Fc_d$ of the loss in excess of its irreducible part equals $\tail_d(\bar M)$, where $\bar M$ is the $w_t$-weighted second moment of $m_U$ about its weighted mean, pooled over tokens and times.
\end{lemma}
\begin{proof}
Write the network output as $Wh_i+\mu_0$. The loss in excess of the irreducible part is $\E_{t,i}[w_t\|Wh_i+\mu_0-m_U\|^2]$. Since $h_i$ is an arbitrary function of the whole input, for fixed $W$ with orthonormal columns and fixed $\mu_0$ the best choice is $Wh_i=P_W(m_U-\mu_0)$; this uses that $m_U$ conditions on the whole grid, the same input as $h_i$. This leaves $\tr\bigl((I-P_W)\,\E[w_t(m_U-\mu_0)(m_U-\mu_0)^\top]\bigr)$. Minimising over $\mu_0$ gives the weighted mean; minimising over rank-$d$ projections $P_W$ keeps the top $d$ eigendirections (Eckart--Young), leaving $\tail_d(\bar M)$. If the total weight is infinite, as for $w_t=t^{-2}$ with $t$ uniform, apply the argument on $t\in[\delta,1]$: the excess loss over $[0,1]$ is at least that over $[\delta,1]$, which gives lower bounds such as those of \cref{thm:relation-ii}(b), (c) as $\delta\downarrow0$.
\end{proof}

\begin{proof}[Proof of \cref{prop:fiber}]
For each $z$, $\|b+Bf-m^*\|^2$ is minimised exactly when $Bf$ is the orthogonal projection of $m^*-b$ onto $\ran B$, which leaves $(I-\Pi_{\ran B})(m^*-b)$. \emph{(a)} If $B=\gamma_tI$, then $f^*=\gamma_t^{-1}(m^*-b)=\gamma_t^{-1}[(a^*-b)+r^*]$. The two parts are orthogonal because $r^*$ is orthogonal to every affine function (orthogonality principle). \emph{(b)} If $B=P_C$, then $\Pi_{\ran B}=P_C$, and components of $f$ in $\row L$ are annihilated; the minimum-norm solution is $P_C(m^*-b)$.
\end{proof}
\subsection{Relation I}

\begin{proof}[Proof of \cref{thm:relation-i}]
\emph{(a)} The first statement is immediate from \eqref{eq:target}. With $|x|=|\varepsilon|=R$ and $z=(1-t)x+t\varepsilon$, expanding $|z|^2$ gives $x^\top\varepsilon=(|z|^2-\nu_tR^2)/(2t(1-t))$. Substituting into $z^\top x=(1-t)R^2+t\,x^\top\varepsilon$ and using $\nu_t+2t(1-t)=1$ gives \eqref{eq:sphere-h}. Similarly, $z^\top(\varepsilon-x)=(1-2t)(x^\top\varepsilon-R^2)=(2t-1)(R^2-|z|^2)/(2t(1-t))$. If one norm is not fixed, $x^\top\varepsilon$ is not determined by $|z|$; for $n=768$ with Gaussian noise, the radial target deviates from the closed form by 3--17\% of the radial component over $t\in[0.3,0.8]$.
\emph{(b)} The first claim is the conversion $\widehat v=(z-\hat x)/t$. For the second, let $z_s=[\sin((1-s)\Omega)\,x+\sin(s\Omega)\,\varepsilon]/\sin\Omega$ with $|x|=|\varepsilon|=R$. Eliminating $\varepsilon$ from $u^S=\dot z_s$ and using $x^\top z_s=R^2\cos(s\Omega)$ gives
\[
  u^S=\Omega\cot(s\Omega)\,z_s-\frac{\Omega}{\sin(s\Omega)}\,x=-c\,P_T(z_s)\,x,\qquad c=\frac{\Omega}{\sin(s\Omega)} .
\]
Pointwise in $z$, the weighted loss is minimised by $\E[\psi u^S\mid z]/\E[\psi\mid z]$, and since $P_T$ depends only on $z$ this equals $-P_T(z)\,\E_\psi[c\,x\mid z]$; for $\psi\equiv1$ it is $m^S=\E[u^S\mid z]$. By \cref{prop:fiber}(b) with $B=P_T$, the optimal outputs are $\{f:\ P_Tf=-P_T\E_\psi[cx\mid z]\}=\{-\E_\psi[cx\mid z]+\lambda\zh\}$. The term $\Omega\cot(s\Omega)z_s$, which is proportional to $z$, never needs to be output.
\emph{(c)} With $B=\beta_tP_T$, $\ran B$ is the tangent space, so by \cref{prop:fiber} the formulation error is $\E\|\zh\zh^\top(m^*-b)\|^2$. It equals $\E\|\zh\zh^\top m^*\|^2$ for $b=0$, and vanishes for $b=L^\dagger h_m=\zh\zh^\top m^*$, which requires $Lm^*=h_m$ to hold exactly, as in (a). On the geodesic path $u^S$ is tangent, so the radial target is zero.
\end{proof}

\begin{proof}[Proof of \cref{prop:pt-floor}]
By the tower property $\Cov(z,D)=\Cov(z,x)=\alpha_t\Sigma_x$; by the law of total variance $\Cov(D)=\Sigma_x-P_t$; and $\Cov(z)=\alpha_t^2\Sigma_x+\sigma_t^2I$. Expanding $\Cov(\kappa_tz+\beta_tD)$ gives the identity. Since $0\preceq P_t\preceq\Sigma_x$, on the null space of $\Sigma_x$ only $\kappa_t^2\sigma_t^2$ remains, which equals $1$ for velocity prediction on the chord path. The MMSE does not exceed the linear MMSE, $P_t\preceq\Sigma_x-\Cov(\Dlin)$, with equality if and only if $D$ is affine. Hence $\Cov_t(m^*)$ dominates its value for Gaussian data and Gaussian noise with the same covariance, which is diagonal in the eigenbasis of $\Sigma_x$ with entries
\[
  e_v(\lambda,t)=\frac{\bigl(t-(1-t)\lambda\bigr)^2}{(1-t)^2\lambda+t^2}
\]
for velocity prediction. By \cref{lem:eckart} and the monotonicity of $\tail_d$ in the Loewner order, $\Ec_d$ is at least the sum of the $n-d$ smallest values of $\bar e_v(\lambda_j)=\int_0^1e_v(\lambda_j,t)\,dt$. The substitution $u=t/(1-t)$ gives
\[
  \bar e_v(\lambda)=\int_0^\infty\frac{(u-\lambda)^2}{(u^2+\lambda)(1+u)^2}\,du=1+\lambda-\frac{\pi}{2}\sqrt\lambda ,
\]
so this sum is at least $(n-d)\min_\lambda\bar e_v(\lambda)=(1-\pi^2/16)(n-d)$, the minimum being attained at $\lambda=\pi^2/16\approx0.617$. With several token positions the bound persists, because $e_v$ is operator convex in $\Sigma$ and mean differences between positions add a positive semidefinite term. For clean-data prediction the corresponding entry tends to $0$ as $\lambda\to0$.

\emph{Single image.} With one image, $x$ is constant at each position and the network can read the position, so $P_t=0$ and the irreducible error vanishes. The field of velocity prediction then has covariance $\kappa_t^2\sigma_t^2I=I$ across the noise, so $\Ec_d\ge(n-d)\,\E_t[q_t]$ per token, that is $(n-d)/n$ per element with unit weights. With Gaussian noise the target $\varepsilon-x_i$, pooled over positions, has covariance $I+\Cov_i(x_i)$, so when $d$ is at least the rank of $\Cov_i(x_i)$, at most the number of positions minus one, the bound holds with equality: the head spans the deviations $x_i-\bar x$, the bias carries the mean, and the residual is $A\varepsilon$ with $A=I-P_W$ of rank $n-d$. Its radial part is $\zh^\top A\varepsilon=\bigl((1-t)\,x_i^\top A\varepsilon+t\,\varepsilon^\top A\varepsilon\bigr)/|z|$. For $|x_i|=\sqrt n$, $\varepsilon^\top A\varepsilon=n-d+O(\sqrt n)$, $|z|^2=n\nu_t+O(\sqrt n)$, and the first term is $O(\sqrt n)$, so its mean square per element is $q^2t^2/\nu_t+O(1/n)$ with $q=(n-d)/n$; and $\int_0^1(t^2/\nu_t)\,p(t)\,dt=\frac12$ for every time density with $p(t)=p(1-t)$, because $t^2+(1-t)^2=\nu_t$. To leading order in $1/n$, the radial part of the residual is therefore $q^2/2$ per element and the angular part $q-q^2/2$. Masking the radial part of the loss, as the projection does, makes $-x/t$ an optimal output (\cref{prop:fiber}(b)); its rank is at most the number of token positions, so every width in the single-image sweep of \citet[Figure~4]{rjf} can fit the angular loss exactly. %

\emph{Equation \eqref{eq:gain}.} With $\widehat v=(z-\hat x)/t$, $\widehat v-v=-(\hat x-x)/t=\beta_t(\hat x-x)$. Moreover $\Cov_t(m_v)=t^{-2}\bigl[\Cov(z)-2\Cov(z,D)+\Cov(D)\bigr]$ with $\Cov(z)=(1-t)^2\Sigma_x+t^2I$ and $\Cov(z,D)=(1-t)\Sigma_x$. Hence $\Cov_t(m_v)-t^{-2}\Cov_t(D)=I+t^{-2}\bigl[(1-t)^2-2(1-t)\bigr]\Sigma_x=I-\frac{1-t^2}{t^2}\Sigma_x$.
\end{proof}

\subsection{Relation II}

\begin{proof}[Proof of \cref{thm:relation-ii}]%
\emph{(a)} Let $s=\cos\Omega=x^\top\varepsilon/R^2$ and $\rho=2t(1-t)/\nu_t$, so $|z|^2=R^2\nu_t(1+\rho s)$; $s$ has density proportional to $(1-s^2)^{(n-3)/2}$ on $[-1,1]$ whatever the data. Since $z^\top x=h_D$ almost surely, $z^\top D=h_D$ and $P_TD=D-c_{\rm ex}z$, with $c_{\rm ex}=h_D/|z|^2=(1-t+ts)/(\nu_t(1+\rho s))$ by \eqref{eq:sphere-h}. Under (E), $m^*=L^\dagger h_m+\beta_tP_TD$, because $\Pi_Rm^*=L^\dagger h_m$ and $P_Tz=0$, so $f_{\mathrm E}^*=P_TD$. Under (P), the fiber is $\{f:P_Tf=P_TD\}\ni D$. If $x\in S$, the posterior is supported on $S$, so $D\in S$, and $P_{S^\perp}f_{\mathrm E}^*=-c_{\rm ex}P_{S^\perp}z=-c_{\rm ex}tP_{S^\perp}\varepsilon$. Given $x$ and $s$, $\varepsilon-sx$ is uniform on the sphere of radius $R\sqrt{1-s^2}$ in the orthogonal complement of $x$, so $\E[(\omega^\top\varepsilon)^2\mid x,s]=n(1-s^2)/(n-1)$ for a unit vector $\omega\in S^\perp$, with $1-s^2=\sin^2\Omega$. Rotations and reflections of $S^\perp$ preserve the law of $\varepsilon$ given $x$ and leave $s$ unchanged, so the component has zero mean and covariance $t^2e_{\rm ex}(t)I_{n-r}$ with $e_{\rm ex}(t)=\frac{n}{n-1}\E[\sin^2\Omega\,c_{\rm ex}^2]$, which is positive because $1-s^2>0$ and $c_{\rm ex}\ne0$ almost surely.
\emph{(b)} Let $x\in S$. The field error of $f=Wh_i+\mu_0$ under (E) is $\E[q_t\beta_t^2\|f-f_{\mathrm E}^*\|^2]\ge\E[q_t\beta_t^2\|P_{S^\perp}(f-f_{\mathrm E}^*)\|^2]$, so only the directions the data do not occupy are used. By (a), $P_{S^\perp}f_{\mathrm E}^*$ has mean zero and second moment, weighted and pooled over $t$, $e_EI_{n-r}$. The output $P_{S^\perp}(Wh_i+\mu_0)$ uses the head $P_{S^\perp}W$ of rank at most $d$, so the argument of \cref{lem:eckart} in $S^\perp$ gives at least $\tail_d(e_EI_{n-r})=(n-r-d)_+e_{\mathrm E}$. No moment of the target in $S$ enters, and the truncation in the proof of \cref{lem:eckart} covers an infinite total weight or an infinite $e_{\mathrm E}$. For any data on the sphere, imposing the norm constraint on $\widehat m_{\mathrm E}$ gives $L^\dagger h_m+P_T(L^\dagger h_m+\beta_tf)=\widehat m_{\mathrm P}$ for the same $f$, so by \cref{prop:impose} the field error under (P) is at most that under (E) for every $f\in\Fc_d$, and $\Ec_d(\mathrm P)\le\Ec_d(\mathrm E)$. For (P) and for clean-data prediction, take $W$ with columns spanning $S$ ($d\ge r$) and $h_i$ the coordinates of $D$; the field error is then zero. Since $e_{\rm ex}(t)>0$ for $0<t<1$, $e_{\mathrm E}>0$ whenever the time weights are positive on a set of positive measure, and for $r\le d<n-r$ the lower bound for (E) is positive while (P) attains zero. %
\emph{(c)} At $t=1/2$, $\cbar=1$ and $h_D=|z|^2$ by \eqref{eq:sphere-h}, so $L^\dagger h_D=z$. For the bound, with $s$ and $\rho$ as in (a) and using $\nu_t+(1-2t)=2(1-t)^2$,
\[
  h_D-\cbar|z|^2=\frac{(2t-1)\,t}{\nu_t}\,R^2s .
\]
The radial field error of the scalar offload is $\beta_t(h_D-\cbar|z|^2)z/|z|^2$, and by \eqref{eq:sphere-h} the exact radial component of the velocity field is $h_mz/|z|^2$ with $h_m=(2t-1)R^2(1-s)$. Let $A_\rho=\E[s^2/(1-\rho^2s^2)]$. Because the law of $s$ is symmetric, $\E[s^2/(1+\rho s)]=A_\rho$ and $\E[(1-s)^2/(1+\rho s)]=1+(1+\rho)^2A_\rho$. With $\beta_t^2=t^{-2}$ and $1+\rho=1/\nu_t$,
\[
  \eta^{\rm rad}_t=\frac{\E\bigl[\beta_t^2(h_D-\cbar|z|^2)^2/|z|^2\bigr]}{\E\bigl[h_m^2/|z|^2\bigr]}=\frac{(1+\rho)^2A_\rho}{1+(1+\rho)^2A_\rho},
  \qquad
  \eta_t=\frac{\E\bigl[\beta_t^2(h_D-\cbar|z|^2)^2/|z|^2\bigr]}{2R^2}=\frac{(2t-1)^2}{2\nu_t^3}\,A_\rho ,
\]
where $\eta_t$ is the fraction of $\E|U|^2=2R^2$. Since $0\le\rho\le1$, $A_\rho$ lies between $\E[s^2]=1/n$ and $\E[s^2/(1-s^2)]=1/(n-3)$, both Beta-function identities. As $(1+\rho)^2\le4$ and $x\mapsto x/(1+x)$ is increasing, $\eta^{\rm rad}_t\le(4/(n-3))/(1+4/(n-3))=4/(n+1)$; the bound is approached as $t\to1/2$, where $\rho\to1$. Finally $(2t-1)^2/\nu_t^3=8u^2/(1+u^2)^3$ with $u=2t-1$, maximised at $u^2=1/2$ with value $32/27$; hence $\eta_t\le16/(27(n-3))$, below $1/n$ for $n\ge8$. Under (S), $B$ is invertible and $f_{\mathrm S}^*=(m^*-b_{\mathrm S})/\beta_t=D-\cbar z$, which differs from $P_TD$ by the radial term $(c_{\rm ex}-\cbar)z$, small by the bound above. If $x\in S$, $P_{S^\perp}(D-\cbar z)=-\cbar P_{S^\perp}z=-\cbar tP_{S^\perp}\varepsilon$, with covariance $\cbar^{\,2}t^2I_{n-r}$ because $\Cov(\varepsilon)=I$, and the argument of (b) gives $(n-r-d)_+e_{\mathrm S}$. For $e_{\mathrm S}$, the substitution $u=2t-1$ above gives $\nu_t=(1+u^2)/2$, $\cbar=(1-u)/(1+u^2)$ and
\[
  \int_0^1\cbar^{\,2}\,dt=\tfrac12\int_{-1}^{1}\frac{(1-u)^2}{(1+u^2)^2}\,du=\tfrac12\int_{-1}^{1}\frac{du}{1+u^2}=\frac{\pi}{4},
\]
the odd part $-2u/(1+u^2)^2$ integrating to zero. For $e_{\mathrm E}=\int_0^1e_{\rm ex}(t)\,dt$, fix $\Omega$ and denote by $\theta$ the geodesic time of \cref{lem:chord-slerp}(a), so that $\hat x^\top\zh=\cos(\theta\Omega)$ and $c_{\rm ex}=z^\top x/|z|^2=R\cos(\theta\Omega)/|z|$; by \cref{lem:chord-slerp}(c), $dt=\Omega|z|^2\,d\theta/(R^2\sin\Omega)$. Hence
\[
  \sin^2\Omega\int_0^1c_{\rm ex}^2\,dt=\Omega\sin\Omega\int_0^1\cos^2(\theta\Omega)\,d\theta=\sin\Omega\Bigl(\frac{\Omega}{2}+\frac{\sin2\Omega}{4}\Bigr).
\]
Because the law of $s$ is symmetric, $\Omega$ and $\pi-\Omega$ have the same law, and the average of the two values is $\frac{\pi}{4}\sin\Omega$. So $e_{\mathrm E}=\frac{\pi}{4}\frac{n}{n-1}\E\sin\Omega$, and $(n-1)/n=\E\sin^2\Omega\le\E\sin\Omega\le1$ gives the bounds.
\end{proof}

\begin{proof}[Proof of \cref{prop:affine-constraints}]%
\emph{(a)} Since $Lx=c$ almost surely, $L\E x=c$ and $L\Sigma_xL^\top=\Cov(Lx)=0$; any distribution with the same mean and covariance therefore has $\E Lx=c$ and $\Cov(Lx)=0$, hence $Lx=c$ almost surely. The relation itself is $Lv=\kappa_tLz+\beta_tLx=\kappa_tLz+\beta_tc$. The LMMSE denoiser is $\Dlin=\E x+\Cov(x,z)\Cov(z)^{-1}(z-\E z)$, and $L\Cov(x,z)=\Cov(Lx,z)=0$, so $L\Dlin=c$ and $La^*=\kappa_tLz+\beta_tc=h$.
\emph{(b)} Let $f=Wh_i+\mu_0\in\Fc_d$. Because $P_C$ is fixed, $P_Cf=(P_CW)h_i+P_C\mu_0$ is again in $\Fc_d$, and $b+\gamma_tP_Cf$ is the field of $(b,\gamma_tP_C)$ at $f$ and of $(b,\gamma_tI)$ at $P_Cf$; hence $\Ec_d(b,\gamma_tI)\le\Ec_d(b,\gamma_tP_C)$. Conversely, imposing the constraint on $(b,\gamma_tI)$ as in \cref{prop:impose} gives $b^C=L^\dagger h+P_Cb=b$, because $\Pi_Rb=L^\dagger h$, and $B^C=\gamma_tP_C$, with a field error at most that of $(b,\gamma_tI)$ for every $f$; hence $\Ec_d(b,\gamma_tP_C)\le\Ec_d(b,\gamma_tI)$.
\end{proof}

\begin{proof}[Proof of \cref{prop:affine-tail}]%
\emph{(a)} By \cref{prop:fiber}(a) the network target is $f^*=\gamma_t^{-1}(m^*-b)$, and $\Ec_d=\inf_{f\in\Fc_d}\E[q_t\gamma_t^2\|f-f^*\|^2]$. \Cref{lem:eckart} with weights $q_t\gamma_t^2$ gives $\tail_d(\bar M)$, where $\bar M=\E[q_t\gamma_t^2(f^*-\mu)(f^*-\mu)^\top]$. Now $\gamma_t(f^*-\mu)=(a^*-b-\gamma_t\mu)+r^*$. For each $t$ the first term is affine in $z$ and $r^*$ is orthogonal to affine functions of $z$ componentwise, pooled over positions (\cref{app:notation}), so the cross terms vanish and $\bar M=\Abar(b)+\Rbar$.
\emph{(b)} $r^*$ depends on neither $b$ nor $\gamma_t$. Second moments of affine functions of $z$ depend only on the first two moments of $z$, hence on those of $Q$ and on the path. $\Abar(b)$ involves $\gamma_t$ through $\mu$.
\emph{(c)} For $b=a^*$ the target is $\gamma_t^{-1}r^*$, whose mean is zero, so $\Abar=0$. Since $\Abar(b)\succeq0$ and $\tail_d$ is monotone in the Loewner order, no $b$ does better than $\tail_d(\Rbar)$. If $\E_t[q_t\gamma_t^2]=\infty$, as for (S) and clean-data prediction under the velocity loss, (a) is not asserted, but the lower bound in (c) still holds whenever the total weight on each $[\delta,1]$ is finite: the truncation in the proof of \cref{lem:eckart} gives $\Ec_d\ge\tail_d(\Rbar_\delta)$, with $\Rbar_\delta$ computed from the weights $q_t\mathbf 1\{t\ge\delta\}$, and $\tail_d(\Rbar_\delta)\to\tail_d(\Rbar)$ as $\delta\downarrow0$. %
\emph{(d)} $\tail_d$ of an $n\times n$ matrix vanishes for $d\ge n$.
\end{proof}

\begin{lemma}[Bounds in the affine family]\label{lem:affine-bounds}
In the setting of \cref{prop:affine-tail},
$\tail_d(\Rbar)+\tail_d(\Abar(b))\le\Ec_d(b,\gamma_tI)\le\tail_d(\Rbar)+\sum_{j\le n-d}\lambda_j(\Abar(b))$.
\end{lemma}
\begin{proof}
By \cref{prop:affine-tail}(a), $\Ec_d=\tail_d(\Abar+\Rbar)$. Use $\tail_d(M)=\min_P\tr(PMP)$ over orthogonal projections of rank $n-d$. The minimum of a sum is at least the sum of minima, which gives the lower bound. For the upper bound, take the $P$ that minimises $\tr(P\Rbar P)$ and apply the Ky Fan inequality $\tr(P\Abar P)\le\sum_{j\le n-d}\lambda_j(\Abar)$.
\end{proof}

\emph{Equation \eqref{eq:trace}.} The minimiser of $\E\|x-cz\|^2$ is $c=\E[x^\top z]/\E|z|^2=(1-t)\E|x|^2/((1-t)^2\E|x|^2+t^2n)$, and $\E|x|^2=\tr\Sigma_x$ for centred data. %
 In the variables of \citet{edm}, $x_{\rm EDM}=z/(1-t)$ and $\sigma=t/(1-t)$; minimising $c_{\rm out}^2=(1-c_{\rm skip})^2\sigma_{\rm data}^2+c_{\rm skip}^2\sigma^2$ gives the same coefficient with $\sigma_{\rm data}^2=\bar\lambda$.

\subsection{Relation III}

\begin{proof}[Proof of \cref{thm:relation-iii}]%
$\Pi_Rr^*=L^\dagger Lm^*-L^\dagger La^*=L^\dagger(h-La^*)$. Since $\Pi_Rm^*=L^\dagger h$, $m^*-k_0=P_C(m^*-a^*)=P_Cr^*$, and the energy identity is the pointwise decomposition $r^*=\Pi_Rr^*+P_Cr^*$.
\end{proof}

\begin{proof}[Derivation for \cref{ex:two-clusters}]
The LMMSE denoiser is $\Dlin=Gz$ with $G=(1-t)\Sigma_x\Cov(z)^{-1}$, which acts on $S_i$ as $g_i=\frac{(1-t)\lambda_i}{(1-t)^2\lambda_i+t^2}$. On cluster $i$, $x^\top Gx=g_iR^2$ is constant, so $\E[z^\top Gz\mid i]=(1-t)^2g_iR^2+t^2\tr G$, while $\E[h_D\mid i]=(1-t)R^2$ for both clusters; the radial deviation from the constraint, in denoiser units, is
\[
  \E\bigl[z^\top Gz-h_D\mid i\bigr]=(1-t)^2g_iR^2+t^2\tr G-(1-t)R^2 .
\]
 The overall mean of $z^\top Gz$ also equals $(1-t)R^2$, by LMMSE orthogonality, so the two cluster deviations are equal in size and opposite in sign; they are nonzero when $g_1\ne g_2$, that is when $k_1\ne k_2$.
\end{proof}

\section{Chord path and geodesic path}\label{app:chord}

This appendix relates the two paths used with spherical latents. Changing the path changes the Bayes field itself, so it is not a formulation in the sense of \cref{sec:formulations}.

\begin{lemma}\label{lem:chord-slerp}%
Let $z_t=(1-t)x+t\varepsilon$ with $|x|=|\varepsilon|=R$, endpoint angle $\Omega$, and $U^E=\varepsilon-x$. Then:
\begin{enumerate}
\item $\zh_t$ is the point at time $s(t,\Omega)=\frac1\Omega\operatorname{atan2}\bigl(t\sin\Omega,\,1-t+t\cos\Omega\bigr)$ of the geodesic interpolation, on the unit sphere, between the normalised endpoints;
\item $|z_t|=\rho R$ with $\rho^2=1-2t(1-t)(1-\cos\Omega)$, so $\cos\Omega=1-\frac{R^2-|z|^2}{2t(1-t)R^2}$: on the chord, the norm reveals the angle;
\item $\partial_ts=\sin\Omega/(\Omega\rho^2)$;
\item $U^E=k\,z+\frac{\sin\Omega}{\Omega\rho}\,u^S_{s(t,\Omega)}$ with $k=(2t-1)(1-\cos\Omega)/\rho^2$, where $u^S$ is the geodesic velocity at radius $R$.
\end{enumerate}
\end{lemma}

\begin{proposition}\label{prop:chord-slerp}%
For a single token, or for independent tokens, the following holds. For each $t$ the map $z\mapsto(\zh,\Omega)$ is a bijection, and
\[
  m^E(z,t)=k\,z+\frac{\sin\Omega}{\Omega\rho}\,m^S_\Omega\bigl(R\zh,s(t,\Omega)\bigr),\qquad m^S_\Omega(y,s)=\E\bigl[u^S_s\mid z^S_s=y,\Omega\bigr],
\]
where $k$, $\rho$ and $\Omega$ are known functions of $(|z|,t)$. The chord field differs from the geodesic field $m^S(y,s)=\E[u^S_s\mid z^S_s=y]$ by known transformations (A1: the radial term $kz$, the input scale $\rho$, the time reparameterisation $s(t,\Omega)$, and an output scale equivalent to the time weight $\sin^2\Omega/(\Omega^2\rho^2)$) and by a genuinely different posterior (A2: conditioning on each token's $\Omega$ rather than averaging over it, and token-specific times $s_j$).
\end{proposition}

The proof conditions the chord on $(\zh_j,\Omega_j)_j$ and notes that $k$, $\rho$ and $\partial_ts$ depend only on $\Omega$. \Cref{lem:chord-slerp} applies token by token on any grid; the relation between the Bayes fields is stated for a single token or independent tokens, because with dependent tokens both fields condition on the whole grid. The size of A2 depends on the data.

\section{Further results}\label{app:further}

\begin{proposition}[Relation II for a general spectrum, at fixed $t$]\label{prop:general-spectrum}%
In the setting of \cref{sec:relation-ii}, let $\lambda_1\ge\dots\ge\lambda_n$ be the eigenvalues of $\Sigma_x=\Cov(x)$; the mean need not vanish. Let $g(\lambda,t)=\frac{(1-t)\lambda}{(1-t)^2\lambda+t^2}$ and $\bar a(\lambda,t)=\bigl(g(\lambda,t)-\cbar\bigr)^2\bigl((1-t)^2\lambda+t^2\bigr)$. If
\[
  \sum_{j>d}\lambda_j\ <\ \sum_{n-d\ \text{smallest}}\bar a(\lambda_j,t),
\]
then at time $t$ the projection has smaller reducible error than the scalar offload.
\end{proposition}
\begin{proof}
At fixed $t$, $\Cov(z)=(1-t)^2\Sigma_x+t^2I$ and $G=(1-t)\Sigma_x\Cov(z)^{-1}$ are diagonal in the eigenbasis of $\Sigma_x$, so $\Cov((G-\cbar I)z)$ is diagonal with entries $\bar a(\lambda_j,t)$; \cref{lem:affine-bounds} bounds the error of (S) below by $\beta_t^2$ times the sum of the $n-d$ smallest entries. With a nonzero mean $\mu_x$, $\Dlin=\mu_x+G(z-(1-t)\mu_x)$, and the constant part of $\Dlin-\cbar z$ is absorbed by the bias $\mu_0$, so the centred covariance $\Cov((G-\cbar I)z)$ enters as before; $\cbar$ is still fixed by $\E|x|^2=R^2$. For (P), $\|\beta_tP_T(f-D)\|\le|\beta_t|\|f-D\|$, so its error is at most $\beta_t^2\tail_d(\Cov_tD)$ by \cref{lem:eckart}, and $\Cov_tD=\Sigma_x-P_t\preceq\Sigma_x$ with $\Sigma_x$ centred and $\tail_d$ monotone. The per-direction threshold at $t=1/2$ solves $(1-\lambda)^2/(4(1+\lambda))=\lambda$, that is $3\lambda^2+6\lambda-1=0$.
\end{proof}

\begin{figure}[ht] %
\centering
\includegraphics{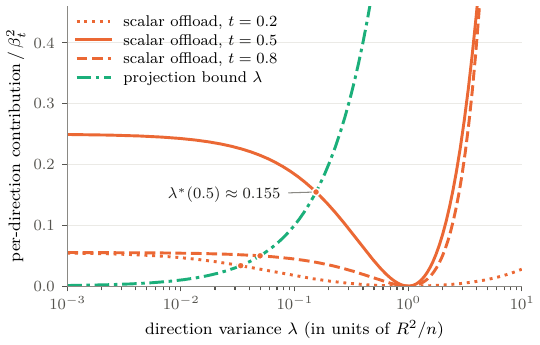}
\caption{Per-direction comparison behind \cref{prop:general-spectrum}: the affine residual $\bar a(\lambda,t)$ of the scalar offload, which bounds its error from below, against the bound $\lambda$ of the projection, as functions of the direction's variance $\lambda$ in units of $R^2/n$, for $t=0.2,0.5,0.8$. Below the crossing $\lambda^*(t)$ (dots) the scalar offload's contribution is the larger. Closed forms.}
\label{fig:general-spectrum}
\end{figure}

The entries $\bar a(\lambda_j,t)$ are the per-direction residuals with which \citet{hybridspectral} decompose an input bottleneck, here for the scalar offload, and the threshold at $t=1/2$ is $\lambda^*=(2\sqrt3-3)/3\approx0.155$. \Cref{fig:general-spectrum} compares the two per-direction contributions. The two sides of the condition sum over differently chosen directions: the left over the $n-d$ smallest variances, the right over the $n-d$ smallest residuals, which vanish at variance $R^2/n$, where the scalar coefficient is exact. Many directions below $\lambda^*$ therefore do not by themselves satisfy the condition. The upper bound for the projection replaces $\Cov_t(D)=\Sigma_x-P_t$ by $\Sigma_x$ and is loose at intermediate and high noise.

\paragraph{Composition of statistics and constraints.} The analytically available component $k_0=L^\dagger h+P_Ca^*$ of \cref{sec:relation-iii} combines the analytic parts of the two kinds of information. Whether it is everything they determine together is a separate question.

\begin{appdefinition}[Jointly determined component]\label{def:components}%
Let $\Wc_\Kc$ be the closure of $\Hc_\aff+\Hc_R$, where $\Hc_R$ is the range of $\Pi_R$. Two fields that satisfy all known relations differ by an element of $\Wc_\Kc^\perp$, so they share the \emph{jointly determined component} $k_\Kc:=\Proj_{\Wc_\Kc}m^*$. Because this projection uses the inner product of $p_t$, which $\Kc$ need not determine, $k_\Kc$ need not be computable from $\Kc$.
\end{appdefinition}

With only statistics $k_\Kc=k_0=a^*$, and with only a constraint $k_\Kc=k_0=L^\dagger h$.

\begin{proposition}[Composition]\label{prop:composition}%
\leavevmode
\begin{enumerate}
\item $k_0=k_\Kc$ if and only if $k_0$ also satisfies the moment relation, $\Pi_\aff k_0=a^*$, that is, $\Pi_\aff L^\dagger(h-La^*)=0$.
\item The criterion in (a) holds under overlap, and it holds when $\Pi_R$ and $\Pi_\aff$ commute; then also $m^*=L^\dagger h+P_Ca^*+P_Cr^*$, a sum of three mutually orthogonal terms contributed by the constraint, the statistics and the data.
\end{enumerate}
\end{proposition}
\begin{proof}
\emph{(a)} Because $\Hc_\aff\subseteq\Wc_\Kc$, $\Pi_\aff k_\Kc=\Pi_\aff m^*=a^*$, which gives the necessity of the criterion. Conversely, $k_0-a^*=L^\dagger h-\Pi_Ra^*=L^\dagger(h-La^*)$. If $\Pi_\aff k_0=a^*$, then $m^*-k_0=P_Cr^*$ is orthogonal to $\Hc_R$ pointwise and to $\Hc_\aff$ because $\Pi_\aff(m^*-k_0)=a^*-\Pi_\aff k_0=0$; it is therefore orthogonal to $\Wc_\Kc$, which contains $k_0$, so $k_0=k_\Kc$. When the criterion fails, $k_0\ne k_\Kc$, and since $k_\Kc$ is the unique nearest point of $\Wc_\Kc$ to $m^*$, $\|m^*-k_\Kc\|<\|m^*-k_0\|$.
\emph{(b)} Under overlap $L^\dagger(h-La^*)=0$. For commuting projections, $\Proj_{\Wc}=\Pi_\aff+\Pi_R-\Pi_\aff\Pi_R$, so $k_\Kc=a^*+L^\dagger h-\Pi_R a^*=k_0$. The term $L^\dagger h$ is pointwise orthogonal to the other two. Under commutation $P_Ca^*=\Pi_\aff P_Cm^*$ is affine while $P_Cr^*=(I-\Pi_\aff)P_Cm^*$ is orthogonal to affine functions, so these two are orthogonal as well.
\end{proof}

\begin{proposition}[When the projections commute]\label{prop:commute}%
$\Pi_R$ and $\Pi_\aff$ commute if and only if $\Pi_R$ maps affine functions to affine functions. On the sphere, if the law of $z$ has support containing a nonempty open set, $\Pi_R(Az+c)$ is affine if and only if the symmetric part of $A$ is a multiple of $I$ and $c=0$; the projections therefore generally do not commute.
\end{proposition}
\begin{proof}
Both maps are orthogonal projections. If $\Pi_R$ preserves $\Hc_\aff$, it is self-adjoint and so also preserves its orthogonal complement, hence commutes with $\Pi_\aff$. Conversely, commuting gives $\Pi_R\Hc_\aff=\Pi_R\Pi_\aff\Hc=\Pi_\aff\Pi_R\Hc\subseteq\Hc_\aff$. On the sphere, $\Pi_R(Az+c)=\frac{z^\top Az+z^\top c}{|z|^2}z$. Suppose it equals $Mz+d$ on an open set, and write $z=r\omega$ with $r>0$ and $|\omega|=1$. Comparing the terms of order $r$ and $r^0$ along rays gives $(\omega^\top A\omega)\omega=M\omega$ and $(\omega^\top c)\omega=d$ for an open set of directions $\omega$. Hence $c=d=0$, every such direction is an eigenvector of $M$, so $M=\alpha I$, and the quadratic form of $A-\alpha I$ vanishes on an open set of directions, so the symmetric part of $A$ is $\alpha I$.
\end{proof}

\begin{proposition}[Non-commuting projections on the sphere]\label{prop:noncommute}%
\leavevmode
\begin{enumerate}
\item On the sphere at $t=1/2$, $\Pi_\aff k_0\ne a^*$, and hence $k_0\ne k_\Kc$, whenever $\Pr(La^*\ne h)>0$.
\item In general $k_\Kc=L^\dagger h+P_C\tilde a$, where the affine function $\tilde a$ solves a finite-dimensional system of normal equations whose coefficients involve moments of $z$ beyond the second, such as $\E[z_jz_kz_lz_m/|z|^2]$. %
\end{enumerate}
\end{proposition}
\begin{proof}
\emph{(a)} In denoiser units $h_D=|z|^2$. Let $\delta=h_D-z^\top\Dlin$ and $e=\delta z/|z|^2$, so that $k_0=a^*+\beta_te$. LMMSE orthogonality gives $\E[z^\top(D-\Dlin)]=0$, hence $\E\delta=0$, and
\[
  \langle e,\Dlin\rangle=\E\Bigl[\frac{\delta(|z|^2-\delta)}{|z|^2}\Bigr]=-\E\Bigl[\frac{\delta^2}{|z|^2}\Bigr]<0
\]
whenever $\Pr(\delta\ne0)>0$. This expectation is finite because $\delta^2/|z|^2\le|D-\Dlin|^2$. So $e$ is not orthogonal to the affine function $\Dlin$, $\Pi_\aff k_0\ne a^*$, and $k_0\ne k_\Kc$ by \cref{prop:composition}(a).
\emph{(b)} For affine $u$ and $v\in\Hc_R$, $u+v=(\Pi_Ru+v)+P_Cu$, so $\Hc_\aff+\Hc_R=\Hc_R\oplus P_C\Hc_\aff$, an orthogonal sum of a closed and a finite-dimensional subspace, hence closed and equal to $\Wc_\Kc$. Thus $k_\Kc=L^\dagger h+P_C\tilde a$, where $\tilde a$ satisfies $\langle P_C\tilde a,P_Cu\rangle=\langle a^*-L^\dagger h,u\rangle$ for every affine $u$; $P_C\tilde a$ is unique.
\end{proof}

\begin{proposition}[Upper bound for the projected family]\label{prop:projection-bound}%
On the geodesic path with the unweighted loss ($\psi\equiv1$ in \cref{thm:relation-i}(b)), the reducible error of $\{P_Tf:f\in\Fc_d\}$ is at most $\tail_d(\bar M_{g^*})$, the bound of \cref{lem:eckart} for a head predicting $g^*=-\E[c\,x\mid z]$, provided $\E[q_sc_s^2]<\infty$ or $\operatorname{rank}\Sigma_x\le d$. Without this condition the bound can be infinite, because $c\approx1/s$ near the data end, even though the actual reducible error is finite since $|u^S|=\Omega R$. If the data lie in an $r$-dimensional subspace $S$ and $d\ge r$, then for every weight $\psi$ the projected family attains the minimiser $m^S_\psi=-P_T\E_\psi[c\,x\mid z]$ of the weighted loss, $\inf_{f\in\Fc_d}\E\bigl[\psi\,\|P_Tf-m^S_\psi\|^2\bigr]=0$; for $\psi\equiv1$ this is zero reducible error.
\end{proposition}
\begin{proof}
$\|P_Tf-m^S\|=\|P_T(f-g^*)\|\le\|f-g^*\|$. For the last claim, $g^*_\psi=-\E_\psi[c\,x\mid z]$ is a weighted average of points of $S$, so it lies in $S$; a head with $W$ spanning $S$ and $h_i$ the coordinates of $g^*_\psi$ gives $P_Tf=m^S_\psi$.
\end{proof}

The matrix $\bar M_{g^*}$ contains no isotropic pass-through component.

\begin{proposition}[Ranking reversal between formulations]\label{prop:reversal}%
Let $\mathbf Q_1$ and $\mathbf Q_2$ be grid distributions with token marginals $Q_1$ and $Q_2$, let $b$ be an affine offload with $B=\gamma_tI$, and write $\Abar_b(Q_i)$ and $\Rbar(\mathbf Q_i)$ for the matrices of \cref{prop:affine-tail}. If
\[
  \tail_d(\Rbar(\mathbf Q_1))<\tail_d(\Rbar(\mathbf Q_2))\quad\text{and}\quad
  \tail_d(\Rbar(\mathbf Q_1))+\tail_d(\Abar_b(Q_1))>\tail_d(\Rbar(\mathbf Q_2))+\sum_{j\le n-d}\lambda_j(\Abar_b(Q_2)),
\]
then $\mathbf Q_1$ has the larger reducible error under $b$ but the smaller one under the full LMMSE offload.
\end{proposition}
\begin{proof}
By \cref{lem:affine-bounds},
\[
  \Ec_d(b;\mathbf Q_1)\ge\tail_d(\Rbar(\mathbf Q_1))+\tail_d(\Abar_b(Q_1))>\tail_d(\Rbar(\mathbf Q_2))+\sum_{j\le n-d}\lambda_j(\Abar_b(Q_2))\ge\Ec_d(b;\mathbf Q_2).
\]
By \cref{prop:affine-tail}(c), under the LMMSE offload the errors are $\tail_d(\Rbar(\mathbf Q_i))$, and the first is the smaller.
\end{proof}

\paragraph{A pair with a reversal.} Take independent tokens, so that each grid distribution $\mathbf Q_i$ is determined by its marginal $Q_i$. Use Gaussian noise and the clean-data weighting $q_t=t^2$, and compare clean-data prediction ($b=\kappa_tz$, $\gamma_t=\beta_t$, so $q_t\gamma_t^2=1$) with the LMMSE offload. Let $Q_1=\mathcal N(0,\lambda I_n)$. Then $\Rbar(\mathbf Q_1)=0$ and $\Abar_b(Q_1)=c_\lambda I$, with
\[
  c_\lambda=\int_0^1\frac{(1-t)^2\lambda^2}{(1-t)^2\lambda+t^2}\,dt>0 ,
\]
so the error under clean-data prediction is $(n-d)c_\lambda$. Let $Q_2$ be supported on a $(d+1)$-dimensional subspace $S$, with independent, zero-mean coordinates along an orthonormal basis of $S$, every one of them non-Gaussian, and one of variance $\epsilon$. Then $\Rbar(\mathbf Q_2)$ is positive definite on $S$, so $\tail_d(\Rbar(\mathbf Q_2))>0$; if one coordinate were Gaussian, this tail would be zero. Under clean-data prediction the matrix $\Abar_b+\Rbar=\int_0^1\Cov_t(D)\,dt$ is diagonal on $S$ and its entry for the coordinate of variance $\epsilon$ is at most $\epsilon$, so the error is at most $\epsilon$. For $\epsilon<(n-d)c_\lambda$, $Q_2$ is better under clean-data prediction and $Q_1$ under the LMMSE offload.

\clearpage %
\section{Readings of published results}\label{app:cases}%

\begin{table}[h]
\centering
\scriptsize
\caption{Representative methods by burden source, information and operation. Relation numbers refer to Sections~\ref{sec:relation-i}--\ref{sec:relation-iii}; ``Decoded'' refers to \cref{sec:decoded}. Evidence is as reported in the cited version.}
\label{tab:case-table}
\begin{tabular}{@{}p{2.6cm}p{2.4cm}p{2.0cm}p{2.6cm}p{1.2cm}p{3.6cm}@{}}
\toprule
Method & Burden source & Information & Operation & Relation & Evidence (source)\\
\midrule
Velocity or noise prediction & pass-through left to network & path & none & baseline & \cref{prop:pt-floor}; \citet[Table 2]{rsb}\\
Clean-data prediction & pass-through & path & affine conversion & I & \citet[Table 2a,b]{jit}; \citet[Tables 2, 3]{diffusibility}; \citet[Table 2]{rit}; \citet[Table 3]{rsb}\\
EDM preconditioning & pass-through and scalar linear part & scalar second moment & scalar LMMSE offload (trace criterion) & II & \citet[App.\ B.6]{edm}; \citet[Table 10]{jit}\\
Learned target mixing & pass-through; ambient penalty of time-shared linear learners & path; scalar $k$ learned from samples & global scalar mixing & I & \citet[Sec.\ 4, Thm.\ 3.2, App.\ G, H]{jinwang}\\
Wiener mixture & full affine part & mean and covariance & LMMSE offload (family optimum) & II, III & \citet[Eqs.\ 14--16, Sec.\ 4]{hybridspectral}\\
Direction-dependent targets & pass-through outside chosen directions & principal subspace, or learned weights & direction-dependent recovery map ($B$ not scalar) & I & \citet[Sec.\ 4]{asymflow}; \citet[Secs.\ 4, 5]{anisoflow}\\
Learned skip & pass-through & none (learned) & learned affine skip & I & \citet[App.\ B.8]{rsb}\\
Residual widening & unchanged & --- & more bandwidth & I & \citet[Tables 3, 4]{rsb}\\
Wider network or head & unchanged & --- & more width & I & \citet{rae}; \citet[Table 1]{rjf}\\
Tangent projection & pass-through and all radial output & norm constraint & projection; geodesic path; exponential map & I, II, III & \citet[Fig.\ 4, Tables 3, 5]{rjf}\\
Spherical latents & radial supervision (pass-through at the noise end) & imposed norm constraint & projection of output and target; geodesic path & I; Decoded & \citet[Tables 2, 7, 9; Figs.\ 4, 7]{aligning}\\
Endpoint-induced velocity & pass-through (not stated by the authors) & norm constraint & endpoint prediction with log map; fixed speed & I; Decoded & \citet[Sec.\ 3.5]{speech}\\
Normal contraction & none & estimated tangent space & analytic normal contraction & Decoded & \citet[Thms.\ 1--3]{msfm}\\
REPA head & not the main output & --- & auxiliary clean-latent target & auxiliary & \citet[Sec.\ 2.3, Table 17]{raev2}\\
\bottomrule
\end{tabular}
\end{table}

\begin{table}[h]
\centering
\small
\caption{Published results with width $d$ at least the token dimension $n$ (\cref{sec:disc-beyond}). Numbers are FID as reported. Each row compares within one source; data, models, training length and evaluation differ across rows.}
\label{tab:beyond}
\begin{tabularx}{\linewidth}{@{}>{\raggedright\arraybackslash}p{2.7cm}>{\raggedright\arraybackslash}p{3.1cm}>{\raggedright\arraybackslash}X>{\raggedright\arraybackslash}X@{}}
\toprule
Source & $d$ versus $n$ & Comparison & Reported result\\
\midrule
\citet[Table 3]{rsb} & DINOv2-B, $d=n=768$ & plain $\to$ 4-stream residual & velocity $16.56\to3.98$; clean $5.50\to3.93$\\
\citet[Table 2a]{jit} & pixels, $d=n=768$ & clean vs.\ velocity & $8.62$ vs.\ $96.53$\\
\citet[Table 2b]{jit} & $d=768$, $n=48$ & nine target--loss pairs & all in $3.46$--$6.20$\\
\citet[Tables 2, 3]{diffusibility} & $d=1536$, $n=1024$ / $768$ & clean vs.\ velocity & COCO $16.80$ vs.\ $29.97$; $17.24$ vs.\ $22.20$\\
\citet[Table 2]{rit} & $d=1152$, $n=384$ & clean vs.\ velocity & $2.63$ vs.\ $3.17$ (80 epochs)\\
\citet[Table 1]{rjf} & $n=768$; $d=768,1024,1152$ & method vs.\ Euclidean & gains $17.4$, $2.1$, $0.66$\\
\citet[Table 7]{aligning} & FLUX.2, $n=128$, $d=768$ & without vs.\ with tangent projection & $25.79$ vs.\ $20.55$\\
\citet[Section 5.1]{jinwang} & VA-VAE, $n=32$, $d=1152$ & learned $k$ vs.\ velocity target, 64 epochs & $2.05$ vs.\ $2.08$\\
\bottomrule
\end{tabularx}
\end{table}

\paragraph{Two further instances of Relation I.} \Citet[Section 3.5]{speech} write the spherical velocity as $\frac{1}{1-t}\log_{x_t}(\hat x_1)$ from a predicted endpoint, in their time convention, and note that on the sphere clean-data and velocity prediction differ less than in Euclidean space. This is the structure of \cref{thm:relation-i}(b). \Citet{aligning} measure the radial share of the linear-path velocity in VAE latents. At the noise end the radial component of $\varepsilon-x$ is the pass-through term, so its energy share is about $r_\varepsilon^2/(r_\varepsilon^2+r_x^2)$ for nearly orthogonal endpoints: about $50\%$ for equal radii, and $89\%$ for the radii $1.38$ and $3.94$ of REPA-E FLUX.1, against the roughly $90\%$ they report. %

\begin{table}[h]
\centering
\footnotesize
\caption{Instances of the distinction between learning difficulty and decoded consequence. ``Decoder'' instances concern $J_G$; ``sampling'' instances concern $\Phi$ or the trajectory. Evidence types: E, empirical result reported in the cited work; T, theorem in the cited work; I, our reading.}
\label{tab:decoded}
\begin{tabularx}{\linewidth}{@{}l>{\raggedright\arraybackslash}p{3.3cm}>{\raggedright\arraybackslash}X l@{}}
\toprule
Class & Instance & Finding & Evidence\\
\midrule
Decoder & \citet{rjf}, inference radius & known radial component; FID $7.79\to6.77$ when the radius moves from $27.7$ to $45$ & E\\
Decoder & \citet{aligning}, swap tests & VAE decoders: replacing only the radius changes the image by LPIPS $0.04$--$0.18$, only the direction by $0.88$--$0.93$ & E\\%
Decoder & \citet{aligning}, decoder swap & flow and decoder must share support: FID $112.17$ and $43.17$ for the two mismatched pairings & E\\%
Decoder & \citet{diffusibility}, single image & artefacts under velocity prediction with a reconstruction-fine-tuned decoder & E, I\\
Sampling & \citet{msfm} & analytic normal contraction bounds distance to the manifold & T, E\\
Sampling & \citet{speech} & speed fixed to $\pi R/2$ instead of the Bayes-optimal magnitude, to avoid short trajectories & I\\
\bottomrule
\end{tabularx}
\end{table}

\section{Numerical checks}\label{app:checks}

We checked the following numerically (Python standard library, seed $0$, under a minute each).
\begin{itemize}
\item Chord--geodesic correspondence, the radial closed form, and the deviation under Gaussian noise (\cref{thm:relation-i}(a), \cref{lem:chord-slerp}): correspondence error $1.7\times10^{-15}$.
\item The field-covariance identity on non-Gaussian atomic data (\cref{prop:pt-floor}): relative error at most $0.6\%$ with $1.5\times10^5$ samples per $t$; and the closed form of $\bar e_v(\lambda)$, which agrees with quadrature to $10^{-13}$ at six values of $\lambda$.
\item Single image (\cref{prop:pt-floor}): for $n=768$, a head spanning $d$ coordinates and a token mean with half its energy outside them, the radial and angular parts of the rank-optimal residual are $0.1255$ and $0.3742$ at $d=384$ and $0.0563$ and $0.2776$ at $d=512$ by Monte Carlo, against $q^2/2$ and $q-q^2/2$.
\item \Cref{thm:relation-ii}(c): for $n=768$, $\max_t\eta_t=7.72\times10^{-4}$ against the upper bound $7.75\times10^{-4}$. \Cref{ex:two-clusters}: the cluster deviations, $\pm1.9\%$, $\pm19\%$ and $\pm33\%$ of $\E h_D$ at $t=0.2$, $0.5$ and $0.8$ ($n=768$), agree with Monte Carlo; the identity in the proof of \cref{prop:noncommute}(a).
\item \Cref{thm:relation-ii}(c): the $S^\perp$ covariance block equals $\cbar^{\,2}t^2I$ within $1.5\%$ ($n=16$, $r=3$, $6\times10^4$ samples); the preconditioners of EDM and \citet{jit} equal the scalar LMMSE to $10^{-15}$; the thresholds $\lambda^*(t)$ of \cref{prop:general-spectrum}. For (E), the closed form in the proof of \cref{thm:relation-ii}(c) gives $e_{\mathrm E}=0.785910$ at $n=768$, which direct double quadrature over $t$ and $s$ reproduces; the diagonal mean of the $S^\perp$ block of the second moment of $P_TD$ (\cref{thm:relation-ii}(a)) equals $t^2e_{\rm ex}(t)$ within $0.3\%$, with off-diagonal entries below $2.1\times10^{-3}$ and mean below $5\times10^{-3}$ (as above, at $t=0.3$, $0.5$, $0.8$).
\end{itemize}

\section{Representation latents}\label{app:latents}

\paragraph{Protocol.} We use the frozen encoders of representation autoencoders \citep{rae}: DINOv2-B with registers (input $224$), SigLIP2-B/16 and MAE-B/16 (input $256$), each giving $256$ tokens of dimension $n=768$. As in \citet{rae}, the final LayerNorm is applied without its affine parameters, so every token has norm $\sqrt n$ up to the LayerNorm $\epsilon$ (relative deviation below $10^{-6}$); we project each token exactly onto the sphere of radius $\sqrt n$. A second variant first applies the per-channel, per-position standardisation of \citet{rae}, with the statistics they release, and then projects each token onto the same sphere. The images are two per class from the ImageNet-1k training set (seed $0$), centre-cropped to $256$ and resized to the encoder's input size. The mean and covariance of the token marginal are estimated from the $2.56\times10^5$ tokens of one image per class, and the Monte Carlo estimates use the tokens of the other; noise is uniform on the sphere of radius $\sqrt n$, the path is the chord and the target the velocity, as in \cref{sec:relation-ii}. Eigenvalues are in units of $R^2/n$. \Cref{prop:general-spectrum} is evaluated with the centred eigenvalues and $\cbar$ from $\E|x|^2=R^2$, for $d\in\{128,256,384,512,640\}$ and $t=0.05,0.10,\dots,0.95$. The complementary part of \cref{thm:relation-iii} uses $a^*=\kappa_tz+\beta_t\Dlin$ with the LMMSE $\Dlin$ from the mean and covariance, and is estimated from $2\times10^4$ samples per $t$. On a CPU the computation takes about fifteen minutes.

\makeatletter\setlength{\@fptop}{0pt}\makeatother
\begin{table}[ht]
\centering
\footnotesize
\caption{Relation III on representation latents. $\lvert\mu_x\rvert^2/n$: energy share of the mean; $\chi_t=\E\|L^\dagger(h-La^*)\|^2/(2R^2)$: energy of the complementary part of \cref{thm:relation-iii} as a fraction of the target energy $\E|U|^2=2R^2$, with its maximum over $t$ in units of $1/n$, the $t$ attaining it, the maximum as a percentage, and the range of $t$ on which it exceeds $1/n$ by more than two standard errors.}
\label{tab:latents}
\begin{tabularx}{\linewidth}{@{}l c c c c >{\raggedright\arraybackslash}X@{}}
\toprule
Encoder & $\lvert\mu_x\rvert^2/n$ & $n\max_t\chi_t$ & at $t$ & $\max_t\chi_t$ & $\chi_t>1/n$ for $t\in$\\
\midrule
\multicolumn{6}{@{}l}{\emph{Encoder outputs, without standardisation}}\\
DINOv2 & $0.213$ & $4.26$ & $0.40$ & $0.55\%$ & $[0.15,0.70]$\\ %
SigLIP2 & $0.271$ & $8.22$ & $0.40$ & $1.07\%$ & $[0.10,0.75]$\\ %
MAE & $0.764$ & $2.97$ & $0.30$ & $0.39\%$ & $[0.10,0.55]$\\ %
\midrule
\multicolumn{6}{@{}l}{\emph{After the standardisation of \citet{rae}, projected to the sphere}}\\
DINOv2 & $0.206$ & $3.21$ & $0.35$ & $0.42\%$ & $[0.15,0.60]$\\ %
SigLIP2 & $0.0003$ & $5.14$ & $0.40$ & $0.67\%$ & $[0.15,0.65]$\\ %
MAE & $0.021$ & $1.75$ & $0.40$ & $0.23\%$ & $[0.20,0.55]$\\ %
\midrule
\multicolumn{6}{@{}l}{\emph{Scalar second moment (closed form, any data on the sphere)}}\\
--- & --- & $0.593$ & $0.15$, $0.85$ & $0.077\%$ & none\\ %
\bottomrule
\end{tabularx}
\end{table}

\begin{figure}[ht]
\centering
\includegraphics[width=\linewidth]{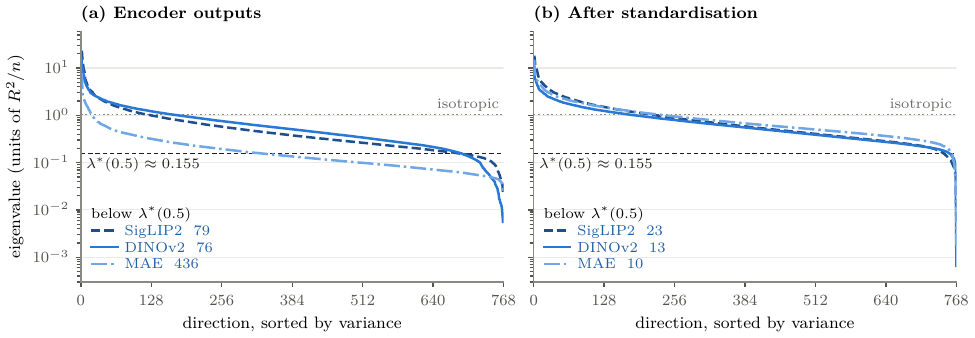}
\caption{Spectra of the representation latents of this appendix: eigenvalues of the token covariance in units of $R^2/n$, sorted, for (a) the encoder outputs and (b) the outputs after the standardisation of \citet{rae}, each projected onto the sphere of radius $\sqrt n$. Dashed: $\lambda^*(0.5)=(2\sqrt3-3)/3$, below which a direction contributes more to the lower bound for the scalar offload than to the upper bound for the projection (\cref{prop:general-spectrum}); the legend gives the number of eigenvalues below it. Dotted: isotropic data. The widths $d$ of the protocol are the ticks from $128$ to $640$. In (a) the smallest eigenvalue, about $5\times10^{-14}$, lies below the axis: the LayerNorm without affine parameters removes the mean over channels, so every token is orthogonal to the all-ones vector.}
\label{fig:latent-spectra}
\end{figure}

\paragraph{Reading.} The complementary part exceeds the average energy per direction at intermediate $t$ for all three encoders and both variants. It does not come from the mean: for SigLIP2 the standardised mean is negligible and the complementary part still peaks at $5.1/n$. The condition of \cref{prop:general-spectrum} holds only for the MAE outputs without standardisation, whose mean carries $76\%$ of the energy against $2\%$ after the standardisation: at $t$ from $0.45$ to $0.60$ for $256\le d\le512$ and from $0.50$ to $0.55$ for $d=128$, with the ratio of its right to its left side reaching $1.27$. In the other five cases this ratio stays below $0.19$, although $10$ to $79$ directions lie below $\lambda^*(0.5)=0.155$ (\cref{fig:latent-spectra}), so the criterion does not decide the comparison of (S) with (P) there. The Monte Carlo relative standard errors, which cover the sampling of noise and tokens for the fixed image pool and the fitted moments, are below $1.4\%$; the tail $\sum_{j>d}\lambda_j$ of the centred covariance, the left side of the condition, differs by at most $1.8\%$ between the two halves of the images. %

\paragraph{Feature definition.} With the final LayerNorm affine restored, the $R_{90}$ of MAE (the number of principal components of the $L^2$-normalised, mean-subtracted tokens that explain $90\%$ of the variance) is $106$, close to the $103$ reported by \citet{diffusibility}; without it, as here, it is $514$ (DINOv2: $520$ and $486$; SigLIP2: $461$ and $475$). %

\section{Decoder sensitivity}\label{app:decoder}

\paragraph{Protocol.} We use the ViT-XL decoders released with representation autoencoders \citep{rae}: for DINOv2-B, decoders trained with additive Gaussian noise on their input of strength $\tau\in\{0,0.5,0.8,1.0\}$ \citep[Section~4.3]{rae}, and for SigLIP2-B and MAE-B the released decoders with $\tau=0.8$. The decoder input is the encoder output of \cref{app:latents} without standardisation, so every token has norm $\sqrt n$ and is orthogonal to the all-ones vector. The latents are those of $250$ images of \cref{app:latents}, the second image of every fourth class; the principal directions are the eigenvectors of the token covariance of \cref{app:latents}, estimated from the first images. For each token $z_i$ the probe $u_i$ is $\pm\zh_i$ (radius), $\pm\mathbf 1/\sqrt n$ (all-ones) or $T_iP_bg_i$, where $g_i\sim\mathcal N(0,I_n)$, $T_i=I-\zh_i\zh_i^\top$ and $P_b$ projects onto one of twelve bands of principal directions, $64$ per band from the highest variance down ($63$ in the last, after removing the all-ones eigenvector). All $256$ tokens are perturbed at once with independent probes, scaled so that the perturbation has mean square $1$ per coordinate, the mean square of the tokens. The gain of a class is the mean of $\big[(G(z+hu)-G(z-hu))/2h\big]^2$ over the $3\times256\times256$ pixel values in $[0,1]$, with $h=0.01$, averaged over four probes per image and over the images; the ratios in \cref{tab:decoder} are ratios of these means, with $95\%$ bootstrap intervals over images. The computation takes about $75$ minutes on one GPU.

\begin{table}[ht]
\centering
\footnotesize
\caption{Decoder gains per unit energy at $h=0.01$. $\rho_{\rm low}$: lowest-variance band over highest-variance band; $\rho_{\rm null}$: all-ones direction over highest-variance band; $\rho_{\rm rad}$: radius over the mean of the twelve bands weighted by their sizes. Brackets: $95\%$ bootstrap intervals over the $250$ images. PSNR: reconstruction from the clean latents, in dB.}
\label{tab:decoder}
\begin{tabular}{@{}l c c c c c@{}}
\toprule
Decoder & $\tau$ & PSNR & $\rho_{\rm low}$ & $\rho_{\rm null}$ & $\rho_{\rm rad}$\\
\midrule
DINOv2-B & $0$ & $19.64$ & $51.067$ \scriptsize$[48.129,54.527]$ & $1.735$ \scriptsize$[1.663,1.813]$ & $0.040$ \scriptsize$[0.035,0.045]$\\ %
DINOv2-B & $0.5$ & $19.08$ & $17.442$ \scriptsize$[16.010,19.003]$ & $0.063$ \scriptsize$[0.056,0.072]$ & $0.237$ \scriptsize$[0.220,0.253]$\\ %
DINOv2-B & $0.8$ & $19.00$ & $15.640$ \scriptsize$[14.523,16.860]$ & $0.049$ \scriptsize$[0.037,0.064]$ & $0.250$ \scriptsize$[0.229,0.272]$\\ %
DINOv2-B & $1.0$ & $18.78$ & $12.721$ \scriptsize$[11.727,13.902]$ & $0.043$ \scriptsize$[0.029,0.063]$ & $0.294$ \scriptsize$[0.272,0.317]$\\ %
SigLIP2-B & $0.8$ & $19.05$ & $42.001$ \scriptsize$[38.897,45.382]$ & $0.023$ \scriptsize$[0.017,0.030]$ & $0.295$ \scriptsize$[0.272,0.317]$\\ %
MAE-B & $0.8$ & $28.19$ & $0.428$ \scriptsize$[0.404,0.451]$ & $0.0018$ \scriptsize$[0.0017,0.0019]$ & $0.069$ \scriptsize$[0.067,0.070]$\\ %
\bottomrule
\end{tabular}
\end{table}

\paragraph{Reading.} For DINOv2-B and SigLIP2-B the gain is nearly flat over the first eleven bands, between $0.57$ and $3.3$ times that of the highest-variance band, and the amplification sits in the last band, whose eigenvalues for DINOv2 reach down to $0.0056$, against $1.7$ to $16$ in the first band (units of $R^2/n$, as in \cref{app:latents}). For MAE-B the highest-variance band is the most sensitive and the others lie at $0.27$ to $0.43$ of it. %
Medians over images, a Gaussian low-pass of the image change and $h=0.04$ instead of $0.01$ leave every ratio on the same side of one, but the ratios depend on the perturbation scale: the last band is the least linear, and its gain relative to the first about halves from $h=0.01$ to $h=0.04$, from $51$ to $27$ for the clean DINOv2-B decoder. %
For some images the decoders have local structures much narrower than $1\%$ of the token scale, where Jacobian--vector products exceed the secant by up to an order of magnitude; we therefore report gains at a stated perturbation scale, and in such images the linearisation~\eqref{eq:decoded} holds only for errors well below it. %

\end{document}